\documentclass[10pt,twocolumn,letterpaper]{article}

\usepackage{cvpr}              

\definecolor{darkgreen}{rgb}{0.0, 0.5, 0.0}

\newcommand{\cmark}{\ding{52}}%
\newcommand{\tbf}[1]{\textbf{#1}}
\newcommand{\ul}[1]{\underline{#1}}
\newcommand{\paragraphhighlight}[1]{\noindent\textbf{#1}\hspace{0.5em}}

\newcommand{\tcr}[1]{\textcolor{red}{#1}}
\newcommand{\tcb}[1]{\textcolor{blue}{#1}}

\definecolor{cvprblue}{rgb}{0.21,0.49,0.74}
\usepackage[pagebackref,breaklinks,colorlinks,allcolors=cvprblue]{hyperref}
\usepackage{xcolor}
\usepackage{array}
\usepackage[table]{xcolor} 
\newcolumntype{g}{>{\columncolor{gray!20}}c}
\usepackage{pifont}
\usepackage{multirow} %
\usepackage{amssymb} 
\def\paperID{} 
\def\confName{CVPR}
\def\confYear{2026}

\title{CycleBEV: Regularizing View Transformation Networks via View Cycle Consistency for Bird’s-Eye-View Semantic Segmentation}

\author{
Jeongbin Hong$^{1,2}$ \quad
Dooseop Choi$^{1,2*}$ \quad
Taeg-Hyun An$^{1}$ \quad
Kyounghwan An$^{1}$ \quad
Kyoung-Wook Min$^{1}$ \\
$^{1}$Electronics and Telecommunications Research Institute (ETRI) \\
$^{2}$University of Science and Technology (UST) \\
{\tt\small \{hjb3880, d1024.choi, tekkeni, mobileguru, kwmin92\}@etri.re.kr }
}

\begin{document}
\maketitle

\def\thefootnote{*}\footnotetext{Corresponding author.}
\def\thefootnote{\arabic{footnote}}

\begin{abstract}
Transforming image features from perspective view (PV) space to bird's-eye-view (BEV) space remains challenging in autonomous driving due to depth ambiguity and occlusion. Although several view transformation (VT) paradigms have been proposed, the challenge still remains. In this paper, we propose a new regularization framework, dubbed CycleBEV, that enhances existing VT models for BEV semantic segmentation. Inspired by cycle consistency, widely used in image distribution modeling, we devise an inverse view transformation (IVT) network that maps BEV segmentation maps back to PV segmentation maps and use it to regularize VT networks during training through cycle consistency losses, enabling them to capture richer semantic and geometric information from input PV images. To further exploit the capacity of the IVT network, we introduce two novel ideas that extend cycle consistency into geometric and representation spaces. We evaluate CycleBEV on four representative VT models covering three major paradigms using the large-scale nuScenes dataset. Experimental results show consistent improvements---with gains of up to 0.74, 4.86, and 3.74 mIoU for drivable area, vehicle, and pedestrian classes, respectively---without increasing inference complexity, since the IVT network is used only during training. The implementation code is available at https://github.com/JeongbinHong/CycleBEV.
\end{abstract}


\section{Introduction}
\label{sec:intro}

Understanding the driving environment is essential for autonomous vehicles (AVs) to navigate safely with reliable recognition. Sensors like LiDAR and radar have been widely used because they provide accurate distance information. However, these sensors are often expensive and do not capture rich details available in camera images, which has driven researchers to focus more on learning unified bird's-eye-view (BEV) representations from surround-view camera images. 
\begin{figure}[t]
\centering
\includegraphics[height=6.5cm]{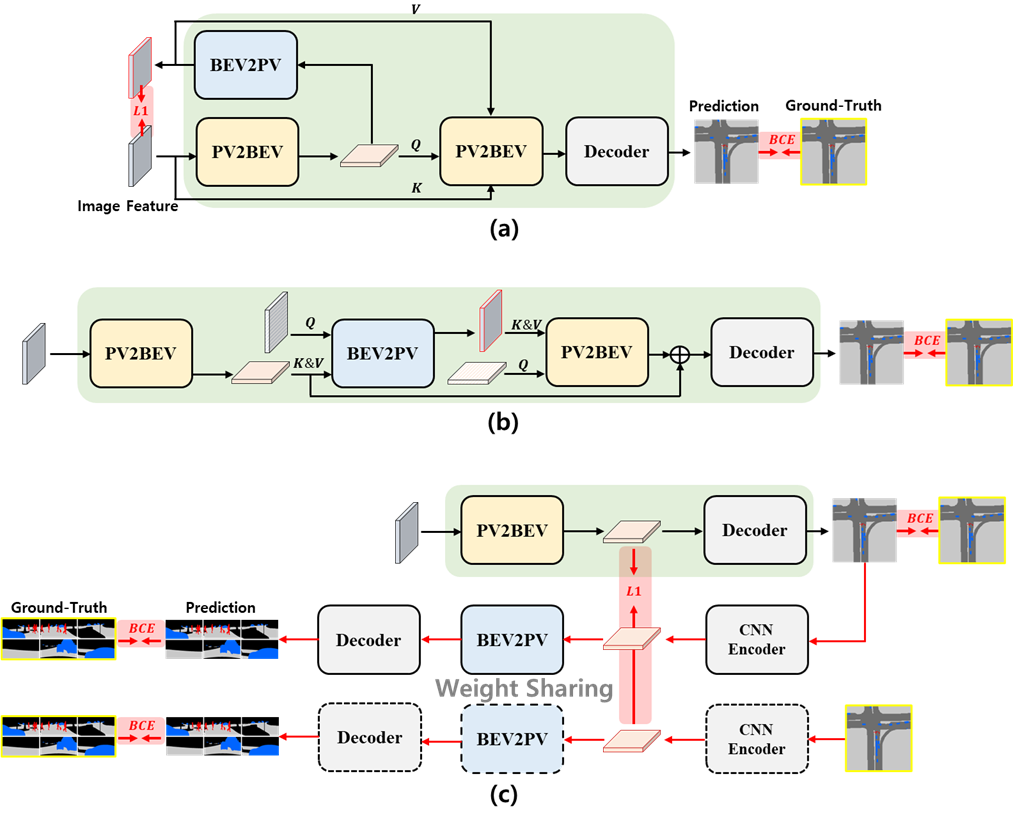}
\caption{\textbf{Visualization of model architectures.} (a) CVTM \cite{Yang_cvpr21}, (b) FocusBEV \cite{Zhao_arxiv24}, and (c) Proposed. Modules outside the green boxes are used only during training. Note that we exclude image feature extraction modules for simplicity. While CVTM and FocusBEV integrate the BEV2PV module, increasing computational cost and network size, the proposed approach employs it only during training. Furthermore, CVTM enforces CC in feature space (semantically vague), and FocusBEV applies feature-space projections without an explicit CC loss (semantically vague and unenforced). As a result, BEV predictions are not directly constrained. In contrast, we enforce semantic-level BEV→PV consistency using a training-only IVT that directly regularizes BEV predictions.}
\label{fig1}
\end{figure}

One of the actively researched tasks that leverage the BEV representations is predicting semantic segmentation maps of driving scenes in BEV space. These maps often include the geometric layouts of driving environments (e.g., drivable space, lane lines) and objects surrounding AVs (e.g., vehicles, pedestrians) in an orthographic BEV perspective. Given that autonomous driving is fundamentally a geometric problem, aimed at ensuring safe and accurate navigation in 3D space, the predicted BEV maps serve as a natural interface for downstream tasks such as motion planning and control. 

The concept of \textit{cycle consistency} was first introduced by \cite{Zhu_iccv17} for unpaired image-to-image translation. They proposed a framework in which a mapping model that translates an image from one domain to another is regularized by enforcing consistency with the reverse mapping, and vice versa. The introduction of a reverse mapping benefits learning a robust mapping because it adds a reversibility constraint that reduces degenerate solutions, regularizes the mapping, and forces the mapping to preserve critical information for reconstruction. Since BEV semantic segmentation is fundamentally a task of learning a mapping from perspective view (PV) to BEV, it is natural to exploit cycle consistency, and several methods have adopted this idea \cite{Yang_cvpr21, Zhao_arxiv24, Monteagudo_wacv25}. However, existing approaches do not fully exploit the potential of the reverse mapping in enforcing cycle consistency as we show in Fig. \ref{fig1}. Moreover, they integrate reverse mapping networks directly into their models, which increases both computational complexity and model size. 

In this paper, we revisit cycle consistency for BEV semantic segmentation. In contrast to the existing methods \cite{Yang_cvpr21, Zhao_arxiv24}, we leverage the reverse mapping capability of an inverse view transformation (IVT) network to regularize the VT network during training, which requires careful IVT design. To this end, inspired by \cite{Swerdlow_ral24}, we devise an IVT network that maps BEV segmentation maps to PV segmentation maps. This IVT network encourages the VT network to capture both richer semantic and geometric information from input PV images. We demonstrate in later sections that our approach consistently improves the performance of the four existing baselines, each representing one of the three major VT paradigms. Finally, to further enrich the proposed cycle-consistency framework, we introduce two novel regularization objectives---\textit{height-aware geometric regularization} and \textit{cross-view latent consistency}---that enhance both geometric and representational coupling between PV and BEV domains. The former extends cycle consistency into geometric space, while the latter extends it into representation space. 



In summary, our contributions are the following.
\begin{itemize}
    \item[$\bullet$] We propose a new regularization framework that effectively leverages view cycle consistency for BEV semantic segmentation, going beyond prior approaches that only partially adopt it.
    \item[$\bullet$] We devise an IVT network that predicts PV segmentation maps from BEV segmentation maps, tailored to our regularization framework, and introduce two novel strategies to further exploit it during training.
    \item[$\bullet$] We apply our framework to four representative baseline models to demonstrate the effectiveness of the proposed approach.
    \item[$\bullet$] We conduct an extensive ablation study and performance analysis to demonstrate the effectiveness of the proposed approach.
\end{itemize}

\section{Related Works}
\label{sec:related_works}

\subsection{View Transformation Paradigms}
Three VT paradigms have dominated over the past five years. The first, commonly referred to as LSS \cite{Philion_eccv20}, estimates per-pixel depth in the input images and combines it with camera parameters to lift image features into 3D space, which are then projected onto BEV grids to obtain BEV features. The second paradigm \cite{Liu_eccv22, Zhou_cvpr22} leverages the cross-attention in Transformer \cite{Vaswani_nips17} to directly learn 2D-to-3D correspondences by querying BEV features from the data. To alleviate the heavy computational cost associated with this approach, the third paradigm \cite{Li_eccv22} introduces deformable cross-attention \cite{Zhu_iclr21} as a more efficient alternative to the standard cross-attention mechanism. Building on these foundations, recent studies have increasingly focused on improving existing paradigms by (1) leveraging temporal consistency in driving scenes \cite{Fang_cvpr23, Liu_iccv23, Xia_eccv24, Chambon_cvpr24}, (2) pre-training 2D image backbones to better capture 3D structure \cite{Yang_cvpr23}, (3) incorporating prior knowledge of ground-truth (GT) BEV maps \cite{Zhu_iccv23, Zhao_cvpr24}, and (4) leveraging the generative capability of Denoising Diffusion Probabilistic Models (DDPMs) \cite{Ji_iccv23, Zou_aaai24, Fu_cvpr25}.

\subsection{View Cycle Consistency}
The concept of view cycle consistency (VCC) was first introduced to BEV semantic segmentation by \cite{Yang_cvpr21}. Specifically, a BEV feature map is derived from a frontal-view (FV) feature map using a VT module, and then used to reconstruct the FV feature map via an IVT module. During training, the reconstructed FV feature map is forced to match the original FV feature map. Building on this idea, \cite{Zhao_arxiv24} proposed reconstructing a FV feature map from the BEV feature map using an IVT module, which is then employed in the VT module as \textit{Key} and \textit{Value} to obtain a denoised BEV representation. Finally, the initial and denoised BEV feature maps are fused for the final prediction. Recently, Monteagudo et al. \cite{Monteagudo_wacv25} proposed a self-supervised BEV segmentation framework that renders frontal-view segmentation maps from predicted BEV features using a pre-trained density field and enforces consistency with frontal-view pseudo-labels. As noted in Sec. \ref{sec:intro}, these methods are limited either by not fully exploiting cycle consistency or by integrating the IVT module into the overall architecture, which results in limited performance gains, increased computational complexity, and network size. Finally, Li et al. \cite{Li_iccv23} first introduced the VCC concept for 3D object detection, in which BEV features whose 3D bounding box predictions fail to correspond to the respective objects in PV images are excluded from subsequent processing. However, this approach is difficult to adapt to BEV semantic segmentation, which considers only 2D grid occupancy.

\subsection{Inverse View Transformation Models}

Regmi et al. \cite{Regmi_cvpr18} first proposed a view transformation framework for reconstructing street-view images from aerial view images using a conditional GAN \cite{Goodfellow_nips14}. Since then, many efforts have focused on aerial-to-street view image synthesis \cite{Toker_cvpr21, Shi_TPAMI22, Li_arxiv24}, while comparatively less attention has been paid to BEV layout-to-PV image synthesis. Swerdlow et al. \cite{Swerdlow_ral24} introduced a view transformer that generates surrounding-view images from a BEV semantic map, and \cite{Wen_cvpr24, Gao_iclr24} extended this idea to controllable video generation. More recently, Li et al. \cite{Li_cvpr25} proposed a unified model that produces surround-view images, LiDAR point clouds, and semantic 3D occupancy grid maps. Inspired by these works, we devise an IVT network to regularize VT networks. Instead of synthesizing realistic PV images—which requires modeling complex image distributions and makes the network cumbersome—it is designed to directly generate PV semantic maps from BEV maps, simplifying the task while retaining essential semantic information.

\section{Proposed Framework}
\label{sec:proposed_framework}

\subsection{Problem Formulation}

Let $I_{i} \in \mathbb{R}^{H_{I} \times W_{I} \times 3}$ denote the image obtained from the $i$-th camera mounted on an AV, where $H_{I}$ and $W_{I}$ respectively denote the height and width of the image. The task is to infer a semantic BEV map $\mathbf{O} \in \mathbb{R}^{X_{o} \times Y_{o} \times |\mathcal{C}|}$, centered at the AV, using the camera inputs $\{ I_{i}\}_{i=1}^{N_{c}}$. Here, $X_{o}$ and $Y_{o}$ denote the map dimensions, $\mathcal{C}$ is the set of semantic categories, and $N_{c}$ is the number of mounted cameras. A typical neural network (NN) pipeline for the BEV prediction proceeds as follows: an image backbone (e.g., ResNet \cite{He_cvpr16}) extracts PV feature maps $ \{\mathbf{F}_{i} \in \mathbb{R}^{H_{F} \times W_{F} \times C_{F}} \}_{i}$ from the input images; a view transformation (VT) encoder then projects these features into a unified BEV representation $\mathbf{B} \in \mathbb{R}^{X_{B} \times Y_{B} \times C_{B}}$; and finally, task-specific heads and decoders operate on $\mathbf{B}$ to generate the prediction $\hat{\mathbf{O}}$. In general, $X_{B} = \lfloor X_{o} / 2^{s} \rfloor$ and $Y_{B} = \lfloor Y_{o} / 2^{s} \rfloor$, where $s$ is zero or a positive integer. 

\subsection{Regularization via View Cycle Consistency}
A conventional way to supervise BEV segmentation models is by minimizing the binary cross-entropy (BCE) loss $\mathcal{L}_{BCE}$ between the ground truth $\mathbf{O}$ and the prediction $\hat{\mathbf{O}}$. Through this supervision, the models learn to map from PV space to BEV space, which we refer to as the \textit{forward mapping}. However, depth ambiguity and occlusion in PV space hinder the models from learning an accurate mapping. To address this issue, we leverage VCC to train VT models by regularizing them with an IVT model that maps BEV space back to PV space. We refer to this as the \textit{reverse mapping}. Through this regularization framework, VT models learn to better capture semantic 3D information from PV images. 


\begin{figure}[t]
\centering
\includegraphics[height=5.5cm]{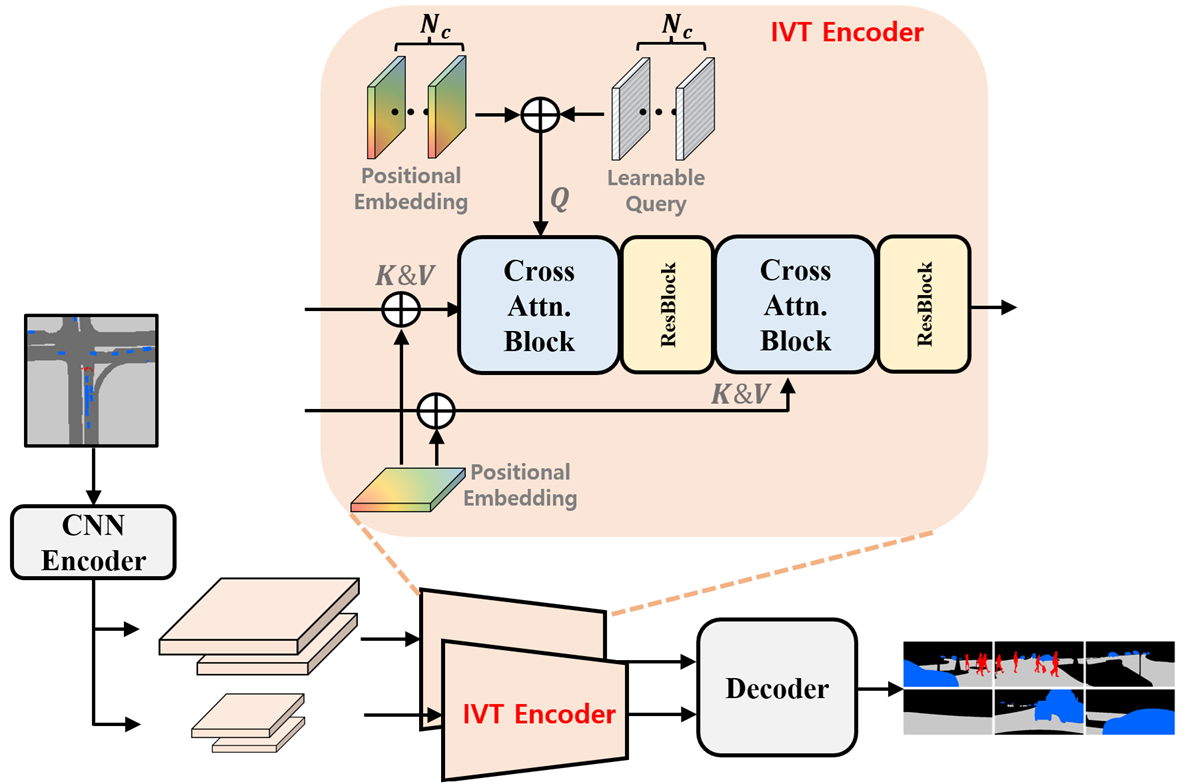}
\caption{\textbf{Illustration of the proposed dual-branch IVT network architecture.}}
\label{fig2}
\end{figure}
\paragraphhighlight{~~IVT Network} Let $\mathcal{F}_{\phi}$ denote the forward mapping function, where $\phi$ represents a set of trainable parameters, indicating that the function can be learned. Given multi-view images $\{ I_{i}\}_{i=1}^{N_{c}}$ as input, $\mathcal{F}_{\phi}$ produces the BEV map $\hat{\mathbf{O}}$ as output. Existing VT models \cite{Philion_eccv20, Liu_eccv22, Zhou_cvpr22, Li_eccv22} are instances of the forward mapping function. It is natural to consider an IVT network that generates multi-view images from BEV maps as the reverse mapping function $\mathcal{R}_{\phi}$. However, this IVT design does not guarantee a one-to-one correspondence between BEV and PV spaces, which may mislead the training of VT models, since reconstructing multi-view images from BEV layouts is inherently ill-posed. To address this issue, we propose an IVT network that generates PV segmentation maps $ \{\mathbf{P}_{i} \in \mathbb{R}^{H_{I} \times W_{I} \times |\mathcal{C}|} \}_{i=1}^{N_{c}}$ from $\mathbf{O}$ with an architecture inspired by \cite{Zhou_cvpr22}. As illustrated in Fig. \ref{fig2}, the proposed IVT network adopts a \textit{dual-branch} design to process multi-resolution (MR) BEV feature maps, which strengthens the regularization of VT models, as discussed in a later section. Specifically, the input BEV map goes through a CNN to become MR BEV feature maps $\{ \bar{\mathbf{B}}_{s} \in \mathbb{R}^{\frac{X_{B}}{2^{s}} \times \frac{Y_{B}}{2^{s}} \times C_{B}} \}_{s}$. Next, $N_{c}$ randomly initialized PV query maps, each corresponding to one of the multi-view PV segmentation maps, attend to the BEV feature maps to update themselves through Transformer's cross-attention. Finally, the updated PV query maps from the two IVT encoders, which are identical in design, are fused and decoded into PV segmentation maps. To help the IVT network learn the mapping from BEV space to PV space, we introduce learnable positional embeddings for the attention based on the following perspective projection equation:
\begin{equation}
\alpha \cdot \mathbf{x}^{I}=\mathbf{K}_{i}\mathbf{R}_{i}(\mathbf{x}^{W}-\mathbf{T}_{i}),
\label{eqn1}
\end{equation}
where $\mathbf{x}^{I}\in \mathbb{R}^{3}$ and $\mathbf{x}^{W}\in \mathbb{R}^{3}$ respectively denote the image pixel and world coordinate points. $\mathbf{K}_{i}$, $\mathbf{R}_{i}$, and $\mathbf{T}_{i}$ represent the intrinsics and extrinsics of the $i$-th camera. For the positional embeddings of $\bar{\mathbf{B}}_{s}$, we first generate grid coordinate points in world space and project them using the right-hand side of Eqn.~\ref{eqn1}. The transformed coordinates are then passed through an MLP before being added to $\bar{\mathbf{B}}_{s}$. On the other hand, grid coordinate points in image pixel space are directly passed through an MLP and added to the PV query maps as the positional embeddings.

\paragraphhighlight{~~Regularization through IVT Network} We regularize VT models under the supervision of the learned reverse mapping function via the following loss:
\begin{equation}
\mathcal{L}_{cycle} = \frac{1}{N_{c}}\sum_{i=1}^{N_{c}}BCE(\mathbf{P}_{i},\hat{\mathbf{P}}_{i}),
\label{eqn2}
\end{equation}
where $\hat{\mathbf{P}}_{i}=\mathcal{R_{\phi}}(\mathcal{F}_{\phi}(\{I_{i}\}_{i}))$ and $\hat{\mathbf{O}}=\mathcal{F}_{\phi}(\{I_{i}\}_{i})$. One may consider the opposite mapping $\mathcal{R}_{\phi} \rightarrow \mathcal{F}_{\phi}$ to further regularize the IVT network during training, following \cite{Zhu_iccv17}. However, VT models take RGB images as input while the IVT produces PV segmentation maps. Therefore, we omit the opposite mapping in our training procedure. Instead, to further enhance BEV segmentation performance, we propose two novel regularization objectives, described in the next section. 

\subsection{Cross-View Regularization Objectives}
\paragraphhighlight{~~Height-Aware Geometric Regularization} As shown in Eqn. \ref{eqn1}, learning the reverse mapping function can be challenging because BEV space lacks the height information (e.g., $\mathbf{x}^{W}=(x^{W}, y^{W},0) \in \mathbb{R}^{3}$), and thus Eqn. \ref{eqn1} no longer holds. To incorporate vertical geometric cues absent in BEV space, we encourage VT models to estimate a height map $\mathbf{H} \in \mathbb{R}^{X_{o} \times Y_{o} \times 1}$ along with $\mathbf{O}$, while the IVT network takes $[\mathbf{H};\mathbf{O}] \in \mathbb{R}^{X_{o} \times Y_{o} \times (|\mathcal{C}|+1)}$ as input. This regularization enforces that the internal geometric representation learned by the VT models remains consistent with the cameras' 3D projection geometry when mapped back to the perspective image planes. To this end, we adapt the decoders of VT models to predict the height map through the following loss:
\begin{equation}
\mathcal{L}_{height} = \frac{1}{L}|| \mathbf{H}-\hat{\mathbf{H}}||_{2}^{2},
\label{eqn3}
\end{equation}
where $\mathbf{H}$ and $\hat{\mathbf{H}}$ denote the GT and predicted height maps, respectively, and $L=X_{o} \times Y_{o}$. Each pixel of $\mathbf{H}$ has a value in $[0,~1]$, representing the normalized height of the moving object occupying the pixel position, while pixels corresponding to road elements are set to 0. Note that a few works \cite{Camiletto_iros24, Park_wacv24} have explored the use of height information. Specifically, \cite{Camiletto_iros24} proposed predicting height from the ground plane using PV images, instead of depth, to accelerate the process of lifting image features into 3D space. \cite{Park_wacv24} predicted a dense height map over a predefined BEV grid, using LIDAR point clouds as supervisory signals, to guide 3D lane detection. To the best of our knowledge, we are the first to encourage VT models to predict the object height map during training, using it as an auxiliary task that serves as a form of regularization. For this purpose, we simply add a dedicated decoder network, which is discarded during inference.

\begin{figure}[t]
\centering
\includegraphics[height=6.7cm]{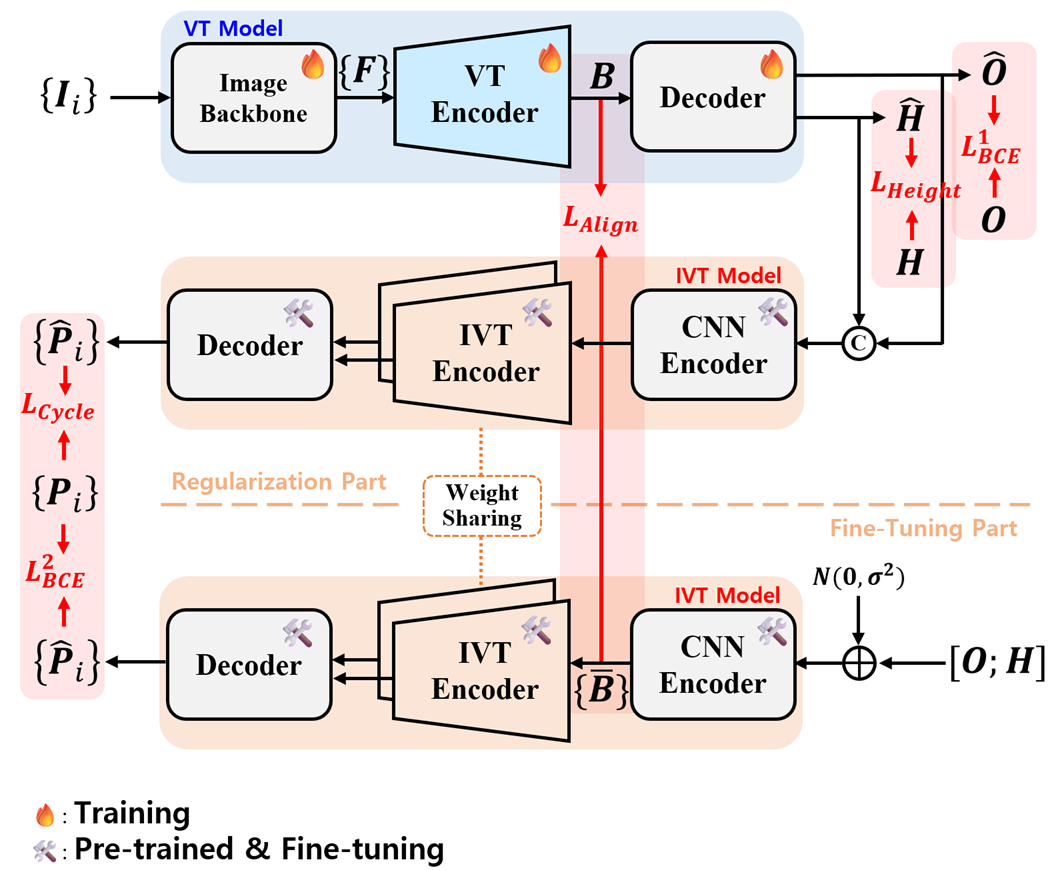}
\caption{\textbf{Visualization of the proposed regularization framework.}}
\label{fig3}
\end{figure}

\paragraphhighlight{~~Cross-View Latent Consistency} MR BEV feature maps $\{ \bar{\mathbf{B}}_{s} \}_{s}$ from the IVT network serve as high-dimensional representations of $[\mathbf{H};\mathbf{O}]$, capturing rich semantic and geometric information about the 3D scene. Therefore, we encourage a BEV feature map, $\mathbf{B}$, from a VT model to share the same representation space with $\{ \bar{\mathbf{B}}_{s} \}_{s}$ through the following alignment loss:
\begin{equation}
\mathcal{L}_{align} = {Smooth}_{\ell_1}(\mathbf{B} - \bar{\mathbf{B}}),
\label{eqn4}
\end{equation}
where ${Smooth}_{\ell_1}$ denotes smooth-$\ell_1$ loss in \cite{Girshick_iccv15}. $\bar{\mathbf{B}}$ is one of $\{ \bar{\mathbf{B}}_{s} \}_{s}$, which has the same size as $\mathbf{B}$ and has undergone FPN \cite{lin_cvpr17}. This alignment enforces latent-space consistency between the VT and IVT networks, effectively coupling their representational geometry across domains. We argue that $\{ \bar{\mathbf{B}}_{s} \}_{s}$, learned through the reverse mapping function, encode richer semantic and geometric cues about the 3D scene than the high-dimensional BEV representations obtained from a conventional BEV map auto-encoder \cite{Zhao_cvpr24}. This argument will be validated in Sec.~\ref{sec:experiments}.

\subsection{Overall Regularization Framework}
Figure \ref{fig3} outlines the proposed regularization framework. Initially, we train the IVT network on pairs of the GT BEV map and its corresponding PV segmentation maps via $\mathcal{L}_{BCE}^{2} = \frac{1}{N_{c}}\sum_{i=1}^{N_{c}}BCE(\mathbf{P}_{i},\hat{\mathbf{P}}_{i})$. Next, we jointly train a VT model and the pre-trained IVT network using the following overall loss:  
\begin{equation}
\begin{split}
\mathcal{L}_{Overall} &= \mathcal{L}_{BCE}^{1} + \lambda_{1}\mathcal{L}_{Height}  \\
&+ \lambda_{2}\mathcal{L}_{Align} + \lambda_{3}\mathcal{L}_{Cycle} + \lambda_{4}\mathcal{L}_{BCE}^{2},
\label{eqn5}
\end{split}
\end{equation}
where $\mathcal{L}_{BCE}^{1}=BCE(\mathbf{O},\hat{\mathbf{O}})$ and $\{\lambda_{i}\}_{i}$ are hyper-parameters. We set $\lambda_{1}=1.0$, $\lambda_{2}=1e^{-3}$, $\lambda_{3}=0.4$, and $\lambda_{4}=1.0$, empirically. When fine-tuning the IVT network (see the bottom part of Fig. \ref{fig3}), we add Gaussian random noise to the input to make it properly handle noisy inputs from the VT network.

\section{Experimental Results}
\label{sec:experiments}

\subsection{Dataset and Evaluation}
We use nuScenes \cite{Caesar} to evaluate our approach. It provides 750, 150, and 150 driving scenes for the training, validation, and test sets, respectively. Since the GT labels for the test set are not available, we use the validation set for evaluation. Following the common practice \cite{Philion_eccv20, Zhou_cvpr22, Choi_iros24, Chambon_cvpr24}, we generate GT BEV segmentation maps at a resolution of $200 \times 200$ by orthographically projecting HD map elements and 3D bounding boxes of vehicles and pedestrians within a $100m \times 100m$ area around the ego-vehicle onto the ground plane. The input camera images are scaled and cropped to $224 \times 448$ pixels before being fed into image backbones. nuScenes partially provides GT PV segmentation maps for multi-view images. Therefore, we first train the SOTA PV segmentation model, Mask2Former \cite{Cheng_cvpr22}, on the available labels. We then use this model to generate pseudo labels for all multi-view images, which serve as training data for the IVT model. For evaluation, we report the Intersection-over-Union (IoU) between the GT and the predictions. All the values reported in the tables in later sections are in mIoU. During training, we consider all objects regardless of their visibility in PV space. Unless otherwise stated, the same setting is applied in evaluation. 


\subsection{VT models and Implementation Details}
We apply our framework to four VT models---LSS \cite{Philion_eccv20}, CVT \cite{Zhou_cvpr22}, PETRv2 \cite{Liu_iccv23}, and BEVFormer \cite{Li_eccv22}---each representing one of the three major VT paradigms. To enable these models to jointly predict the three semantic categories (\textit{drivable area}, \textit{vehicle}, and \textit{pedestrian}) and the height map, we modify CVT's decoder so that the overall decoder architecture includes two sub-decoders---one dedicated to the three categories and the other to the height map---and use this modified decoder for all four models. In addition, we omit the temporal aggregation modules of the baselines, as our focus is on evaluating spatial regularization rather than temporal modeling. We show in the supplementary material how the performance gain varies when incorporating the temporal modules. Except for this modification, we adhere to the original implementations. We also compare our regularization method with those of the existing approaches: CVTM \cite{Yang_cvpr21} and FocusBEV \cite{Zhao_arxiv24}. Specifically, we replace \textbf{BEV2PV} and \textbf{PV2BEV} in Fig. \ref{fig1} with the proposed IVT network and the four VT models, respectively, with minor adjustments to the IVT network. More details can be found in the supplementary material. 


\begin{table}[t]
\begin{center}
\scalebox{0.7}{
\begin{tabular}{|c|c c c|c|}
\hline
\textbf{Model} & \textit{Driv.} & \textit{Veh.} & \textit{Ped.} & Avg. \\
\hline
\textbf{CVT} & 76.80 & 31.41 & 10.89 & 39.70 \\
\textbf{CVT}+\cite{Yang_cvpr21}  & \ul{77.06}$_{0.26\uparrow}$ & \ul{31.61}$_{0.2\uparrow}$ & \ul{11.19}$_{0.3\uparrow}$ & \ul{39.95}$_{0.25\uparrow}$ \\
\textbf{CVT}+\cite{Zhao_arxiv24}  & 76.52$_{0.28\downarrow}$ & 31.23$_{0.18\downarrow}$ & 10.73$_{0.16\downarrow}$ & 39.49$_{0.21\downarrow}$ \\
\textbf{CVT+Ours}  & \tbf{77.40}$_{0.6\uparrow}$ & \tbf{34.24}$_{2.83\uparrow}$ & \tbf{13.69}$_{2.8\uparrow}$ & \tbf{41.78}$_{2.08\uparrow}$ \\
\hline
\textbf{PETRv2} & 78.80 & \ul{31.51} & 8.31 & 39.54 \\
\textbf{PETRv2}+\cite{Yang_cvpr21}  & \ul{79.45}$_{0.65\uparrow}$ & 31.49$_{0.02\downarrow}$ & \ul{8.91}$_{0.6\uparrow}$ & \ul{39.95}$_{0.41\uparrow}$ \\
\textbf{PETRv2}+\cite{Zhao_arxiv24}$^{\dagger}$  & 44.36 &  4.33 & 0.90 & 16.26 \\
\textbf{PETRv2+Ours}  & \tbf{79.54}$_{0.74\uparrow}$ & \tbf{34.25}$_{2.74\uparrow}$ & \tbf{11.74}$_{3.43\uparrow}$ & \tbf{41.84}$_{2.3\uparrow}$ \\
\hline
\textbf{LSS} & 67.58 & 16.85 & \ul{1.34} & 28.59 \\
\textbf{LSS}+\cite{Yang_cvpr21}  & \ul{67.84}$_{0.26\uparrow}$ & \ul{16.88}$_{0.03\uparrow}$ & 1.33$_{0.01\downarrow}$ & \ul{28.68}$_{0.09\uparrow}$ \\
\textbf{LSS}+\cite{Zhao_arxiv24}  & 64.09$_{3.49\downarrow}$ & 10.59$_{6.26\downarrow}$ & 0.98$_{0.36\downarrow}$ & 25.22$_{3.37\downarrow}$ \\
\textbf{LSS+Ours}  & \tbf{67.87}$_{0.29\uparrow}$ & \tbf{21.71}$_{4.86\uparrow}$ & \tbf{5.08}$_{3.74\uparrow}$ & \tbf{31.55}$_{2.96\uparrow}$ \\
\hline
\textbf{BEVFormer} & 78.06 & \ul{33.23} & \ul{11.70} & \ul{41.00} \\
\textbf{BEVFormer}+\cite{Yang_cvpr21}  & 78.10$_{0.04\uparrow}$ & 33.18$_{0.05\downarrow}$ & 11.66$_{0.04\downarrow}$ & 40.98$_{0.02\downarrow}$ \\
\textbf{BEVFormer}+\cite{Zhao_arxiv24}  & \ul{78.17}$_{0.11\uparrow}$ & 32.85$_{0.38\downarrow}$ & 11.22$_{0.48\downarrow}$ & 40.75$_{0.25\downarrow}$ \\
\textbf{BEVFormer+Ours}  & \tbf{78.20}$_{0.14\uparrow}$ & \tbf{34.46}$_{1.23\uparrow}$ & \tbf{13.39}$_{1.69\uparrow}$ & \tbf{42.02}$_{1.02\uparrow}$ \\

\hline
\end{tabular}
}
\end{center}
\caption{\textbf{Quantitative comparison.} The bold and underline indicate the best and second-best performance, respectively. The values in the subscript indicate the performance gain over the baseline. $\dagger$ indicates that the training result could not be made comparable to others despite careful tuning. }
\label{tab1}
\end{table}

\begin{figure*}[t]
\centering
\includegraphics[height=12.0cm, width=17.0cm]{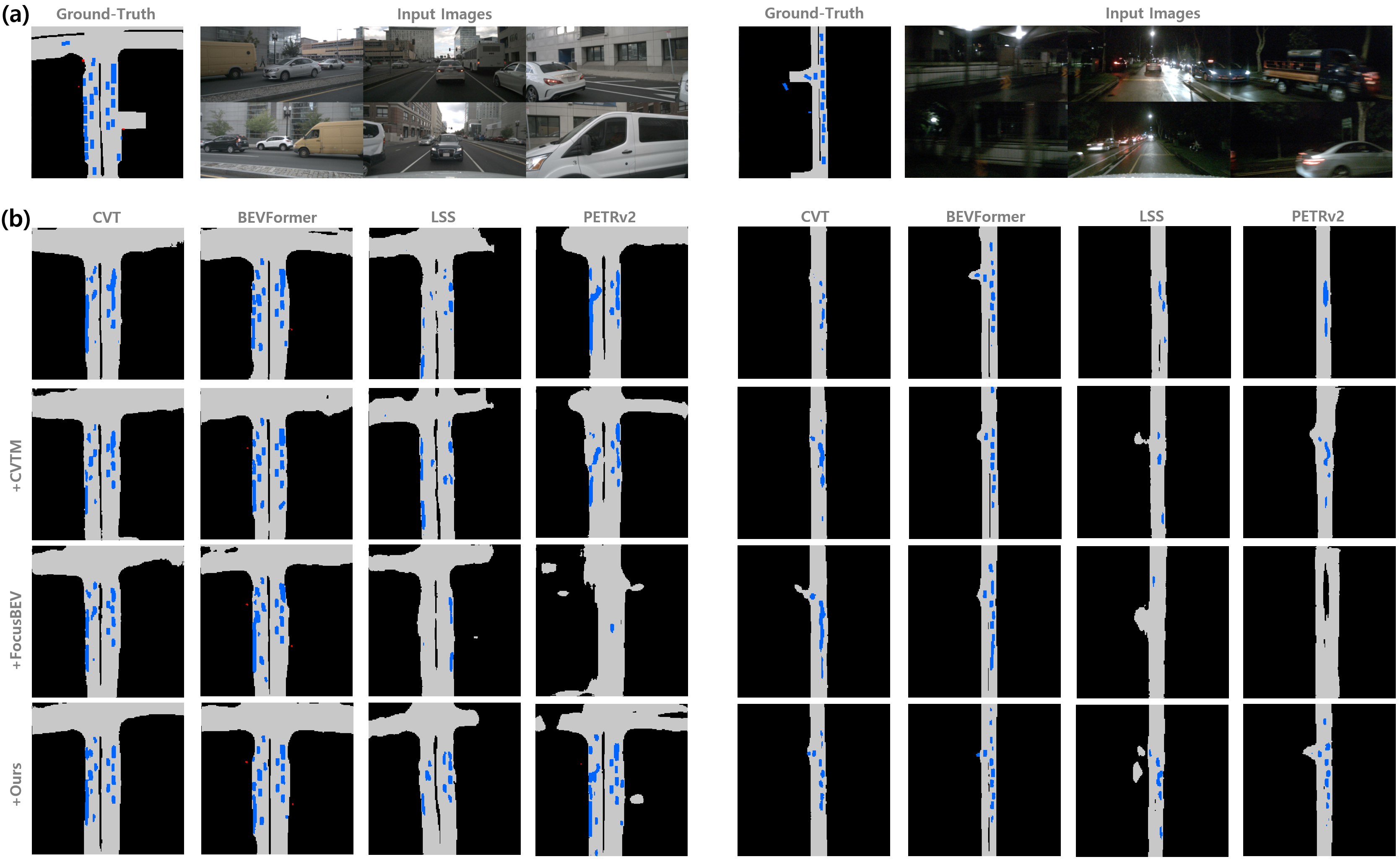}
\caption{\textbf{Prediction results.} (a) Input images and their corresponding ground-truth BEV maps, (b) BEV map prediction results. In (b), the first row shows the predictions from the four baseline models. The second, third, and fourth rows show the results when CVTM \cite{Yang_cvpr21}, FocusBEV \cite{Zhao_arxiv24}, and Ours are applied to the baseline models, respectively. \textit{Drivable area}, \textit{vehicle}, and \textit{pedestrian} are color-coded with gray, blue, and red, respectively. Please zoom in for better visibility.}
\label{fig4}
\end{figure*}
\subsection{Objective Comparison}
In Tab. \ref{tab1}, we quantitatively compare our framework with existing approaches on the four baselines. It shows that the proposed framework consistently improves the performance of the four baselines across all categories, confirming its general applicability to different VT paradigms. In contrast, CVTM \cite{Yang_cvpr21}, which incorporates an IVT network, fails to improve the performance of the baselines across all categories, although it yields average performance gains for most of them, except for BEVFormer. While the proposed method achieves gains of up to 0.74, 4.86, and 3.74 mIoU for \textit{drivable area}, \textit{vehicle}, and \textit{pedestrian}, respectively, CVTM achieves 0.65, 0.2, and 0.6. This is because CVTM makes partial use of VCC. On the other hand, FocusBEV \cite{Zhao_arxiv24}, which integrates an IVT network to exploit VCC implicitly, degrades the performance of all four baselines. The performance drop is particularly noticeable on LSS and PETRv2. We speculate that adopting the IVT network without explicit supervision (e.g., $\mathcal{L}_{cycle}$ in Eqn. \ref{eqn2}) leads to this degradation.

\subsection{Subjective Comparison}
In Fig. \ref{fig4}, we present the input images, the corresponding ground-truth BEV maps, and the prediction results. The figure shows that the proposed framework effectively improves the prediction performance of all four baselines across all categories. In particular, objects that are partially visible in the input images are better predicted with the proposed framework. For example, LSS produces the poorest prediction results in both scenes, failing to detect many partially visible vehicles. However, with the aid of the proposed framework, it successfully detects several of them. In addition, some detections that appear clumped become clearer and more spatially separated. Similar improvements can be observed in other baseline results. On the other hand, CVTM fails to achieve consistent improvements and often predicts vehicles closer together, making them appear more clumped. Similar patterns are even more noticeable in the results of FocusBEV. Additional prediction results are provided in the supplementary materials.


\subsection{Analysis}

\paragraphhighlight{~~Ablation Study} Table \ref{tab2} presents the ablation study of our regularization method using CVT and BEVFormer. The results show that each component consistently contributes to improved performance. Regularizing VT models with the proposed IVT network alone yields noticeable gains, consistent with observations in CVTM. Furthermore, incorporating the two objectives leads to additional noticeable gains. Finally, the IVT network with the dual-branch design outperforms its single-branch counterpart, despite both processing the same number of MR BEV feature maps from the CNN encoder. The main difference between the single- and dual-branch designs is that the former fuses the MR feature maps progressively in one IVT encoder, whereas the latter first processes high and low resolution feature maps separately through dedicated IVT encoders and then fuses the resulting features in the decoding process. As shown in the supplementary material, although the single-branch design performs better than its counterpart in terms of PV segmentation accuracy, the dual-branch design proves more effective at regularizing VT models. Additional details on the comparison between the single- and dual-branch designs are provided in the supplementary material.  

\begin{table}[t]
\begin{center}
\scalebox{0.65}{
\begin{tabular}{|c|c c c|c c c|c|}
\hline
\textbf{Model} & VCC & Height & Align &  \textit{Driv.} & \textit{Veh.} & \textit{Ped.} & Avg.\\
\hline
\multirow{5}{*}{\textbf{CVT}}  
  & &  &                       & 76.80       & 31.41       & 10.89       & 39.70 \\
  & \cmark &  &                 & 76.99       & 32.65       & 12.00       & 40.55  \\
  & \cmark & \cmark &           & 77.12       & 33.76       & 13.32       & 41.40 \\
  & \cmark & \cmark & \cmark  & \tbf{77.40} & \tbf{34.24} & \tbf{13.69} & \tbf{41.78} \\
  & \cellcolor{gray!20}\cmark & \cellcolor{gray!20}\cmark & \cellcolor{gray!20}\cmark  & \cellcolor{gray!20} 77.23 & \cellcolor{gray!20} 34.14 & \cellcolor{gray!20} 13.65 & \cellcolor{gray!20} 41.67 \\
\hline
\multirow{5}{*}{\textbf{BEVFormer}} 
  & &  &                      & 78.06       & 33.23       & 11.70       & 41.00 \\
  & \cmark &  &                 & \tbf{78.20} & 33.61       & 13.32       & 41.71  \\
  & \cmark & \cmark &          & 78.11       & 34.26       & 13.04       & 41.80 \\
  & \cmark & \cmark & \cmark   & \tbf{78.20} & \tbf{34.46} & \tbf{13.39} & \tbf{42.02} \\
  & \cellcolor{gray!20}\cmark & \cellcolor{gray!20}\cmark & \cellcolor{gray!20}\cmark   & \cellcolor{gray!20} 77.76 & \cellcolor{gray!20} 34.31 & \cellcolor{gray!20} 13.34 & \cellcolor{gray!20} 41.80 \\
\hline
\end{tabular}
}
\end{center}
\caption{\textbf{Ablation study on the effectiveness of our contributions.} \textit{VCC}, \textit{Height}, and \textit{Align} refer to view cycle consistency, height-aware geometric regularization, and cross-view latent consistency, respectively. Results highlighted in gray are produced using a single-branch IVT network.}
\label{tab2}
\end{table}

\begin{table}[t]
\begin{center}
\scalebox{0.7}{
\begin{tabular}{|c|c c| c c |}
\hline

\multirow{2}{*}{\textbf{Model}} 
& \multicolumn{2}{c|}{Vis. $\geq$ 40$\%$} &  \multicolumn{2}{c|}{Vis. $<$ 40$\%$} \\ 
&  \textit{Veh.} & \textit{Ped.} & \textit{Veh.} & \textit{Ped.}\\
\hline

\textbf{CVT}              &  33.15      &   11.40     & \ul{8.53}                        & 2.42 \\
\textbf{CVT}+\cite{Yang_cvpr21}              &  \ul{33.43}$_{0.28\uparrow}$      &   \ul{11.74}$_{0.34\uparrow}$     & 8.49$_{0.04\downarrow}$                        & \ul{2.53}$_{0.11\uparrow}$ \\
\textbf{CVT+Ours}         &  \tbf{36.69}$_{3.54\uparrow}$     &   \tbf{14.32}$_{2.92\uparrow}$     & \tbf{9.07}$_{0.54\uparrow}$       & \tbf{2.78}$_{0.36\uparrow}$ \\

\hline

\textbf{BEVFormer}        &  \ul{35.32}      &   12.40     & \ul{9.01}                        & \ul{2.60}\\
\textbf{BEVFormer}+\cite{Yang_cvpr21}        &  35.31$_{0.01\downarrow}$      &   \ul{12.45}$_{0.05\uparrow}$     & 8.90$_{0.11\downarrow}$                        & \ul{2.60}$_{0.0}$\\
\textbf{BEVFormer+Ours}   &  \tbf{36.87}$_{1.55\uparrow}$      &   \tbf{14.31}$_{1.91\uparrow}$     & \tbf{9.42}$_{0.41\uparrow}$       & \tbf{3.00}$_{0.4\uparrow}$\\

\hline

$\Delta_{\mathbf{C-B}}$        &  -2.17     &   -1.0     & -0.48                        & -0.18\\
$\Delta_{\mathbf{CO-B}}$        &  1.37      &   1.92     & 0.06                        & 0.18\\
\hline

\end{tabular}
}
\end{center}
\caption{\textbf{Prediction performance on high and low visible objects.} The values in the subscript indicate the performance gain over the baseline. $\Delta_{\mathbf{C-B}}$ and $\Delta_{\mathbf{CO-B}}$ respectively denote the gains of CVT and CVT+Ours over BEVFormer. }
\label{tab3}
\end{table}

\begin{figure}[t]
\centering
\includegraphics[height=5.75cm]{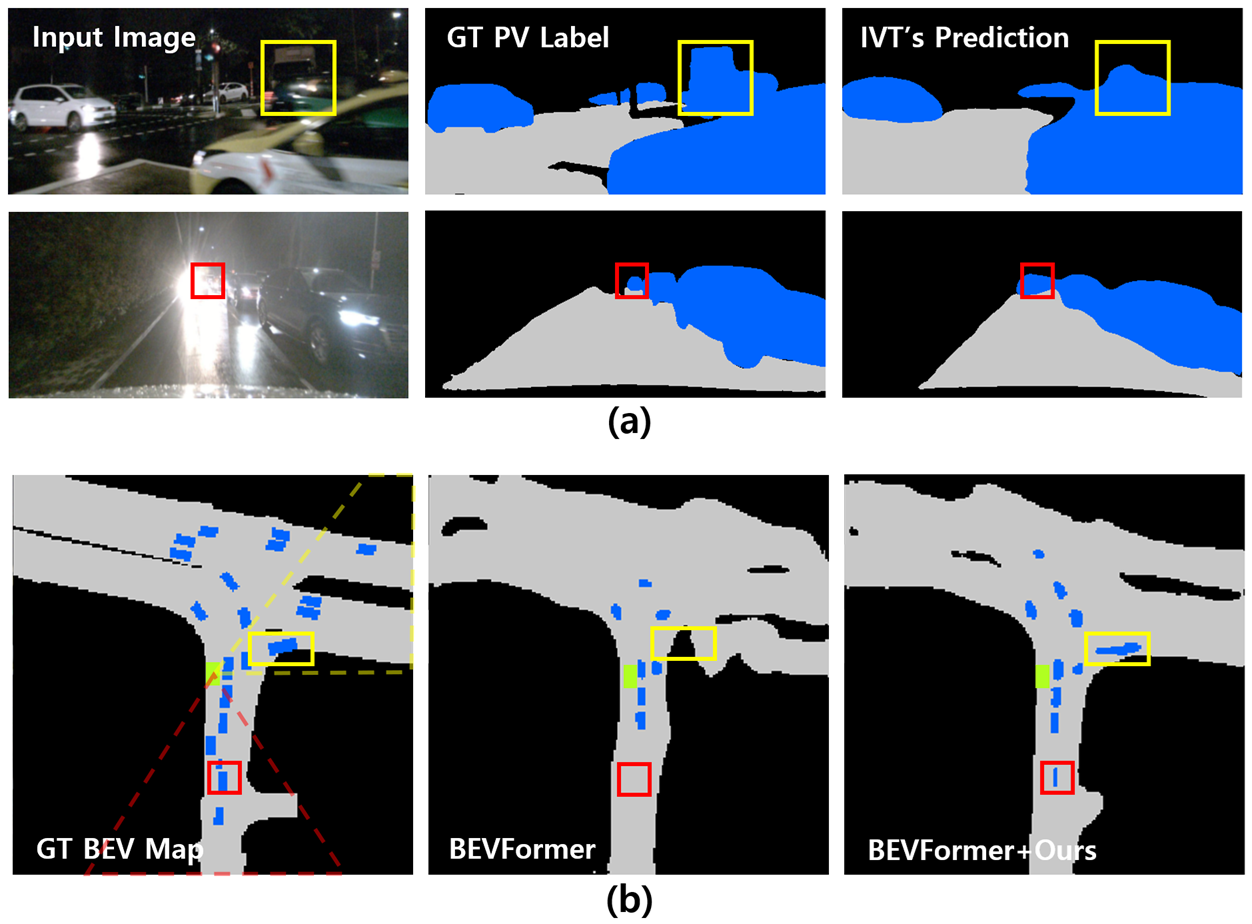}
\caption{\textbf{Prediction examples on a scene with occluded vehicles.} (a) Input images (the first column), ground-truth PV segmentation maps (the second column), and PV segmentation maps predicted by the proposed IVT network (the third column). (b) Ground-truth BEV map (the first column), BEV map predicted by BEVFormer (the second column), and BEV map predicted by BEVFormer+Ours (the third column). The green boxes indicate the AV.}
\label{fig5}
\end{figure}
\paragraphhighlight{~~Robustness to Occlusion} The IVT model learns how objects in BEV space become partially visible in PV space through the reverse mapping. We argue that VT models guided by this capability handle occlusion more effectively, as demonstrated by the results in Tab. \ref{tab3}. The table shows the BEV segmentation performance for objects with visibility greater than 40$\%$ and those with visibility less than 40$\%$, the latter referring to highly occluded objects. It is shown that the proposed method improves the baselines’ ability to handle highly occluded objects, whereas CVTM \cite{Yang_cvpr21} fails. We also report the performance gains of CVT and CVT+Ours over BEVFormer. Since CVT was designed to meet real-time requirements, its original performance is lower than that of BEVFormer, which performs a greater number of cross- and self-attention operations for higher accuracy (see $\Delta_{\mathbf{C-B}}$). However, CVT+Ours achieves higher performance than BEVFormer regardless of visibility level (see $\Delta_{\mathbf{CO-B}}$). Our argument is further supported by the prediction results in Fig.~\ref{fig5}, where partially visible vehicles on the road are more accurately predicted using the proposed framework. Specifically, the vehicles highlighted by the red and yellow boxes in the figure, which BEVFormer fails to detect, is observable in BEVFormer+Ours. Overall, although the IVT network fails to reconstruct the detailed outlines of the vehicles in the PV segmentation map, it still recognizes the presence of low-visibility vehicles, which can lead to improved performance on such objects. 





\begin{table}[t]
\begin{center}
\scalebox{0.7}{
\begin{tabular}{|c|c c c|c|}
\hline
\textbf{Model} & \textit{Driv.} & \textit{Veh.} & \textit{Ped.} & Avg.\\
\hline

\textbf{CVT}  &  76.80      &   31.41     &   10.89     &  39.70 \\
\textbf{CVT+AE}  &  \ul{77.10}$_{0.3\uparrow}$      &   \ul{31.61}$_{0.2\uparrow}$      &   \ul{10.93}$_{0.04\uparrow}$      &  \ul{39.88}$_{0.18\uparrow}$  \\
\textbf{CVT+IVT}  & \tbf{77.40}$_{0.6\uparrow}$        & \tbf{32.05}$_{0.64\uparrow}$        &  \tbf{11.97}$_{1.08\uparrow}$       & \tbf{40.47}$_{0.77\uparrow}$  \\

\hline
\textbf{BEVFormer}   &    78.06     &   33.23     &    11.70    & 41.00 \\
\textbf{BEVFormer+AE}   & 78.06      &  \tbf{33.30}$_{0.07\uparrow}$       &   \ul{11.75}$_{0.05\uparrow}$      & \ul{41.04}$_{0.04\uparrow}$   \\
\textbf{BEVFormer+IVT}   &  \tbf{78.18}$_{0.12\uparrow}$       &   \ul{33.25}$_{0.02\uparrow}$      &   \tbf{11.81}$_{0.11\uparrow}$      & \tbf{41.08}$_{0.08\uparrow}$  \\

\hline
\end{tabular}
}
\end{center}
\caption{\textbf{Comparison with BEV auto-encoder (AE) supervision.} The values in the subscript indicate the performance gain over the baseline.}
\label{tab4}
\end{table}
\paragraphhighlight{~~Comparison with BEV Auto-Encoder Supervision} We argue that the MR BEV feature maps, $\{ \bar{\mathbf{B}}_{s} \}$, generated by the IVT network capture semantic 3D cues more effectively than conventional high-dimensional BEV representations. To validate this claim, we regularize CVT and BEVFormer under two types of guidance using only the loss in Eqn.~\ref{eqn4}: MR BEV feature maps (1) from the pre-trained IVT network and (2) from a pre-trained BEV auto-encoder \cite{Zhao_cvpr24}. Note that the BEV auto-encoder encodes $[\mathbf{H};\mathbf{O}]$ using a series of convolutional layers, whose architecture is nearly identical to that of the IVT network’s CNN encoder, to produce $\{ \bar{\mathbf{B}}_{s} \}$ and then decodes them using another set of convolutional layers. We also freeze the IVT network and the auto-encoder while training the two baselines. Table \ref{tab4} shows the result. The BEV representations from both the IVT model and the auto-encoder provide meaningful supervision signals for the VT models, leading to performance improvements. However, the improvement is more pronounced when the guidance is derived from the IVT network, confirming our claim. 

\begin{table}[t]
\begin{center}
\scalebox{0.7}{
\begin{tabular}{|c|c c c|c|}
\hline
\textbf{Model} & \textit{Driv.} & \textit{Veh.} & \textit{Ped.} & Avg.\\
\hline
\textbf{CVT}  &  \ul{76.80}      &   31.41     &   10.89     &  39.70 \\
\textbf{CVT+SA-I}  &  76.40$_{0.4\downarrow}$      &   \ul{31.75}$_{0.34\uparrow}$    &   \ul{11.43}$_{0.54\uparrow}$    & \ul{39.86}$_{0.16\uparrow}$ \\
\textbf{CVT+SA-II}  &  75.08$_{1.72\downarrow}$      &   31.40$_{0.01\downarrow}$    &   10.92$_{0.03\uparrow}$    & 38.52$_{1.18\downarrow}$ \\
\textbf{CVT+Ours}  & \tbf{77.40}$_{0.6\uparrow}$       & \tbf{34.24}$_{2.83\uparrow}$       & \tbf{13.69}$_{2.8\uparrow}$       & \tbf{41.78}$_{2.08\uparrow}$ \\

\hline
\textbf{BEVFormer}   &    78.06    &   33.23     &    11.70    & 41.00 \\
\textbf{BEVFormer+SA-I}   &    \tbf{78.68}$_{0.62\uparrow}$    &    \ul{33.70}$_{0.47\uparrow}$    &    \ul{12.57}$_{0.87\uparrow}$    & \ul{41.65}$_{0.65\uparrow}$  \\
\textbf{BEVFormer+SA-II}   &    77.11$_{0.95\downarrow}$    &    32.61$_{0.62\downarrow}$    &    11.68$_{0.02\downarrow}$    & 40.47$_{0.53\downarrow}$  \\
\textbf{BEVFormer+Ours}   & \ul{78.20}$_{0.14\uparrow}$       & \tbf{34.46}$_{1.23\uparrow}$       & \tbf{13.39}$_{1.69\uparrow}$       & \tbf{42.02}$_{1.02\uparrow}$ \\

\hline
\end{tabular}
}
\end{center}
\caption{\textbf{Comparison with semantic-aware (SA) image backbone training.} ‘SA-I’: backbones optimized with PV segmentation; ‘SA-II’: backbones optimized with PV segmentation and irrelevant features discarded via the PV segmentation. The values in the subscript indicate the performance gain over the baseline.}
\label{tab5}
\end{table}
\paragraphhighlight{~~Comparison with Semantic-Aware (SA) Image Backbone Training} One may argue that the performance gain of the proposed framework mainly results from involving PV segmentation task during the training of VT models. To demonstrate that the gain is primarily attributable to the IVT’s reverse-mapping capability, we train VT models with an auxiliary task following \cite{Zhang_iccv23}. Specifically, the image backbones of VT models are jointly optimized as part of a PV segmentation model during the training of the VT models (SA-I), and image features irrelevant to target classes are identified and excluded when calculating BEV feature maps (SA-II). Our experiments in Tab. \ref{tab5} show that jointly optimizing the image backbones of the VT models improves the performance of both baselines. However, the gain is smaller than that achieved by the proposed regularization framework. In contrast, excluding image features irrelevant to the target classes degrades performance. This approach, originally designed for 3D object detection, may prevent the VT models from extracting contextual information from the background beneficial to target classes.

\begin{table}[t]
\begin{center}
\scalebox{0.6}{
\begin{tabular}{|c|c c|c c c|c|}
\hline
\textbf{Model} & Aug. &  CycleBEV &  \textit{Driv.} & \textit{Veh.} & \textit{Ped.} & Avg.\\
\hline
\multirow{4}{*}{\textbf{CVT}}  
  &        &        &  76.80       & 31.41       & 10.89       & 39.70 \\
  & \cmark &        &  77.27$_{0.47\uparrow}$       & 31.73$_{0.32\uparrow}$       & 11.08$_{0.19\uparrow}$       & 40.03$_{0.33\uparrow}$ \\
  &        & \cmark &  \ul{77.40}$_{0.6\uparrow}$       & \ul{34.24}$_{2.83\uparrow}$       & \ul{13.69}$_{2.8\uparrow}$       & \ul{41.78}$_{2.08\uparrow}$ \\
  & \cmark & \cmark &  \tbf{77.43}$_{0.63\uparrow}$       & \tbf{34.38}$_{2.97\uparrow}$       & \tbf{13.59}$_{2.7\uparrow}$       & \tbf{41.80}$_{2.10\uparrow}$ \\
\hline
\multirow{4}{*}{\textbf{BEVFormer}} 
  &        &         & 78.06       & 33.23       & 11.70       & 41.00 \\
  & \cmark &         & \ul{79.72}$_{1.66\uparrow}$       & \ul{35.12}$_{1.89\uparrow}$       & \ul{13.40}$_{1.70\uparrow}$       & \ul{42.75}$_{1.75\uparrow}$ \\
  &        & \cmark  & 78.20$_{0.14\uparrow}$       & 34.46$_{1.23\uparrow}$       & 13.39$_{1.69\uparrow}$       & 42.02$_{1.02\uparrow}$ \\
  & \cmark & \cmark  & \tbf{79.73}$_{1.67\uparrow}$       & \tbf{36.38}$_{3.15\uparrow}$       & \tbf{15.19}$_{3.49\uparrow}$       & \tbf{43.77}$_{2.77\uparrow}$ \\
\hline
\end{tabular}
}
\end{center}
\caption{\textbf{Compatibility with typical image augmentation.} The values in the subscript indicate the performance gain over the baseline.  }
\label{tab6}
\end{table}
\paragraphhighlight{~~Compatibility with Data Augmentation} Input data augmentation has been widely used as an effective way to improve model performance without increasing model complexity. Table \ref{tab6} shows the additional gains the proposed framework achieves on top of typical image augmentation. In the experiment, we randomly rotate, resize, and crop input images, and adjust the corresponding camera parameters, following \cite{Fang_cvpr23, Harley_icra23}. The table indicates that the proposed framework is more beneficial to CVT, whereas the image augmentation is more beneficial to BEVFormer when the two approaches are applied individually to the baselines. However, when both approaches are jointly applied, the results surpass those from individual applications, confirming that the proposed framework is compatible with image augmentation. 

\paragraphhighlight{~~Performance Gain under Temporal Setting} We apply the proposed regularization framework to BEVFormer in a temporal setting, where the model is configured to take previously captured images as input. As shown in the supplementary material, the proposed framework improves the performance across all categories over almost all observation horizon lengths. Notably, the static BEVFormer trained under our framework (BEVFormer+Ours in Tab. \ref{tab1}) outperforms the temporal BEVFormer, demonstrating the effectiveness of the proposed approach. More details can be found in the supplementary material. 

\section{Conclusion and Future Work}

In this paper, we presented a new regularization framework, CycleBEV, for BEV semantic segmentation. Viewing the task as learning a mapping function from PV space to BEV space, we proposed learning a reverse mapping function through the proposed IVT network and using it to regularize VT models during training. We introduced two novel regularization objectives to further leverage the reverse mapping capability of the IVT network. Extensive experiments on nuScenes show that CycleBEV consistently improves the four representative VT networks across multiple classes, without increasing inference cost or model size, highlighting its effectiveness and generalizability. Future work will explore extending the proposed framework to temporal VT models that incorporate previously captured images. In particular, we plan to enforce cycle consistency between the current BEV map and earlier frames to enhance temporal coherence. 

\section*{Acknowledgement}
This work was supported by IITP grant funded by the Korea government (MSIT) (RS-2023-00236245, Development of Perception/Planning AI SW for Seamless Autonomous Driving in Adverse Weather/Unstructured Environment)

{
    \small
    \bibliographystyle{ieeenat_fullname}
    \bibliography{main}
}

\clearpage
\setcounter{page}{1}
\maketitlesupplementary


\section{Implementation Details}
In this section, we provide the detailed information about our experiments in Sec. 4. Our implementation code is also provided along with this supplementary material for better understanding. Note that the code will be made public once it is accepted for publication.

\subsection{IVT Network}
\begin{figure*}[t]
\centering
\includegraphics[height=7.0cm]{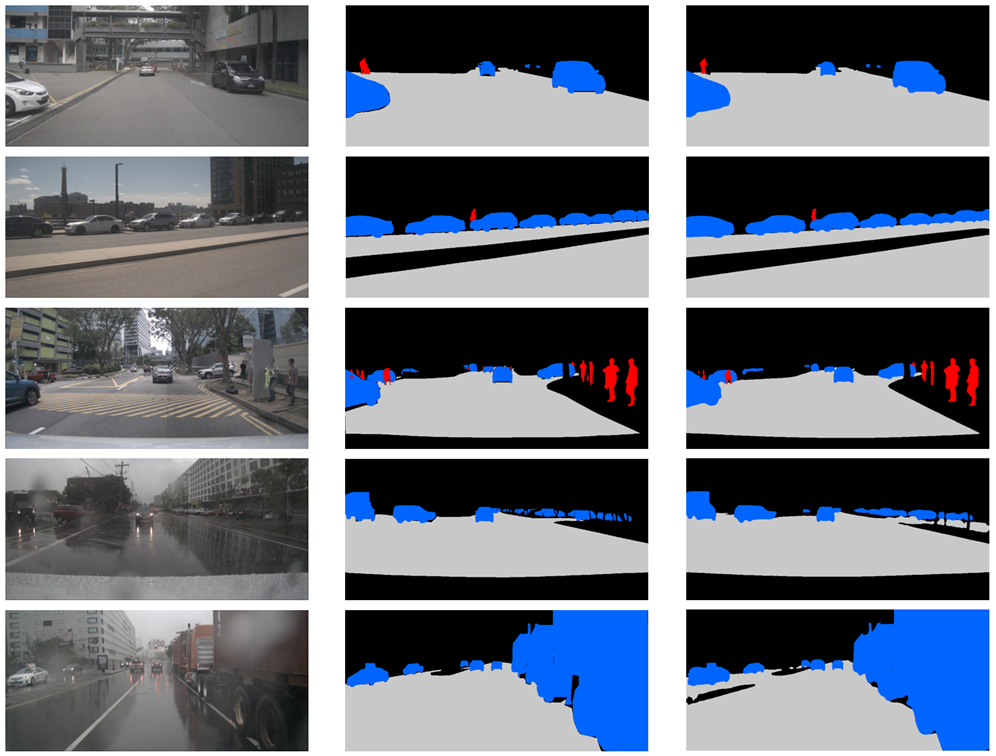}
\caption{\textbf{Visualization of multi-view images (first column), the GT PV segmentation maps (second column), and the pseudo labels generated by Mask2Former (third column).} \textit{Drivable area}, \textit{vehicle}, and \textit{pedestrian} are color-coded with gray, blue, and red, respectively.}
\label{fig4_supple}
\end{figure*}

\paragraphhighlight{~~Pseudo Label Generation} nuScenes partially provides GT PV segmentation maps for
multi-view images, whereas our framework requires GT for all of them. To address this issue, we first train Mask2Former \cite{Cheng_cvpr22} on the available labels. We then use the trained model to generate PV map predictions for all multi-view images in nuScenes, which serve as pseudo labels. Figure \ref{fig4_supple} illustrates examples of GT PV maps and pseudo labels generated by the trained Mask2Former, along with their corresponding input images. From now on, we refer to the pseudo label as the GT label for simplicity.

\paragraphhighlight{~~Pre-training} We train the proposed IVT network on the pairs of GT BEV map and the corresponding GT PV maps using BCE loss. The loss weights for \textit{drivable area}, \textit{vehicle}, and \textit{pedestrian} are empirically set to 0.03, 0.5, and 1, respectively, reflecting the number of pixels in the BEV maps assigned to each category. AdamW \cite{Loshchilov_iclr19} is used for the optimization with an initial learning rate of $4 \cdot 10^{-4}$ and batch size of 4 for 24 epochs.

\paragraphhighlight{~~Fine-tuning} When training VT models under the regularization of the IVT network, we also fine-tune the pre-trained IVT network using AdamW with an initial learning rate of $4 \cdot 10^{-4}$. Specifically, when it receives GT BEV maps, we add random Gaussian noise to the inputs so that the IVT network learns to better handle noisy predictions from the VT models. 

\subsection{Proposed Framework}
\paragraphhighlight{~~Regularization through IVT Network} Given the pre-trained IVT network, we jointly optimize the four baselines---LSS \cite{Philion_eccv20}, CVT \cite{Zhou_cvpr22}, PETRv2 \cite{Liu_iccv23}, and BEVFormer \cite{Li_eccv22}---together with the IVT network under the proposed framework. Note that to enable the baseline models to jointly predict the three semantic categories (\textit{drivable area}, \textit{vehicle}, and \textit{pedestrian}) and the height map, we modify CVT's decoder so that the overall decoder architecture includes two sub-decoders---one for the semantic categories and the other for the height map---and use this modified decoder for all four models. The loss weights for \textit{drivable area}, \textit{vehicle}, and \textit{pedestrian} are empirically set to 0.03, 0.5, and 1, respectively, reflecting the number of pixels in the BEV maps assigned to each category. The hyperparameters in Eqn. \ref{eqn5} are empirically set to $\lambda_{1}=1.0$, $\lambda_{2}=10^{-3}$, $\lambda_{3}=0.4$, and $\lambda_{4}=1.0$. We train the models using AdamW with a batch size 2 for up to 50 epochs. For parameters related to optimization (e.g., initial learning rate), we strictly follow the original implementations of the four baselines. Unless otherwise stated, the same settings are applied to the later experiments. 

\paragraphhighlight{~~Object Height Map Generation} The height map $\mathbf{H}$ in Eqn.~\ref{eqn3} has the same spatial resolution as $\mathbf{O}$, where each of its pixels has a value in $[0, 1]$, representing the normalized height of the moving object occupying that pixel. Pixels corresponding to road elements are set to 0. We obtain height information for moving objects from their ground-truth 3D bounding boxes and normalize it by dividing by 5, which is considered the maximum height (in meters) of a moving object. If the normalized height of a pixel exceeds 1, we clip it to 1. 

\subsection{Existing VCC-based Frameworks}
To implement the VCC-based learning frameworks proposed by CVTM \cite{Yang_cvpr21} and FocusBEV \cite{Zhao_arxiv24}, we let the VT encoders of four baselines and the proposed IVT encoder serve as \textbf{PV2BEV} and \textbf{BEV2PV} in Fig. \ref{fig1}, respectively, with minor adjustments to the IVT network. For CVTM, following the original implementation, we introduce the cycle loss $\mathcal{L}_{cycle}=||\mathbf{X}-\mathbf{X}''||_{1}$, where $\mathbf{X}$ and $\mathbf{X}''$ respectively denote PV image features and those recovered by \textbf{BEV2PV}. In contrast, since FocusBEV does not explicitly enforce cycle consistency via a cycle loss, we optimize it using only the binary cross entropy loss.

\subsection{BEV Map Auto-Encoder}
The BEV autoencoder (AE) used for Tab. \ref{tab4} consists of an encoder and a decoder. The encoder takes the BEV map $\mathbf{O}$ and the height map $\mathbf{H}$ as inputs, while the decoder reconstructs these maps from the high-dimensional representations $\{\mathbf{\bar{B}}_{s} \}_{s}$ produced by the encoder. The encoder comprises a series of \textbf{ResBlock}, each consisting of \textit{Conv}, \textit{BatchNorm}, \textit{MaxPool}, and \textit{ReLU} layers. As the input passes through multiple \textbf{ResBlock}s, the spatial resolution decreases. The decoder has a similar architecture to the encoder, except that (1) it includes multiple skip connections from the encoder, and (2) \textit{DeConv} layers replace \textit{MaxPool} layers. AdamW is used with an initial learning rate of $5 \cdot 10^{-4}$, a batch size of 4, and a total of 10 epochs. During training, we add randomly generated Gaussian noise to $\mathbf{\bar{B}}_{s}$ of the smallest resolution to make the AE robust to noise following \cite{Zhao_cvpr24}.

\subsection{Input Data Augmentation}
We randomly rotate (from $-1^\circ$ to $1^\circ$), resize (scaling from 0.8 to 1.2), and crop (up to 20$\%$ of the image area) the input images to VT models, and adjust the corresponding camera parameters following \cite{Fang_cvpr23, Harley_icra23}. However, we do not modify the corresponding PV segmentation maps accordingly because they are recovered from the BEV maps predicted by the VT models.

\subsection{Semantic-Aware Image Backbone Training}
Following \cite{Zhang_iccv23}, we jointly optimize the image backbones of the four baselines within the PV segmentation task during training. To this end, we devise a PV segmentation decoder, which has a similar architecture to the decoder of UNet \cite{Ronneberger_MICCAI15}. We further apply soft-thresholding to the backbone image features: each feature is scaled by the corresponding PV segmentation logits (after applying a sigmoid for normalization). The resulting scaled features are then used as keys and values in the self- and cross-attention layers.  

\section{Further Analysis}
\subsection{IVT Network Design}
\begin{figure*}[tb]
\centering
\includegraphics[height=7.0cm]{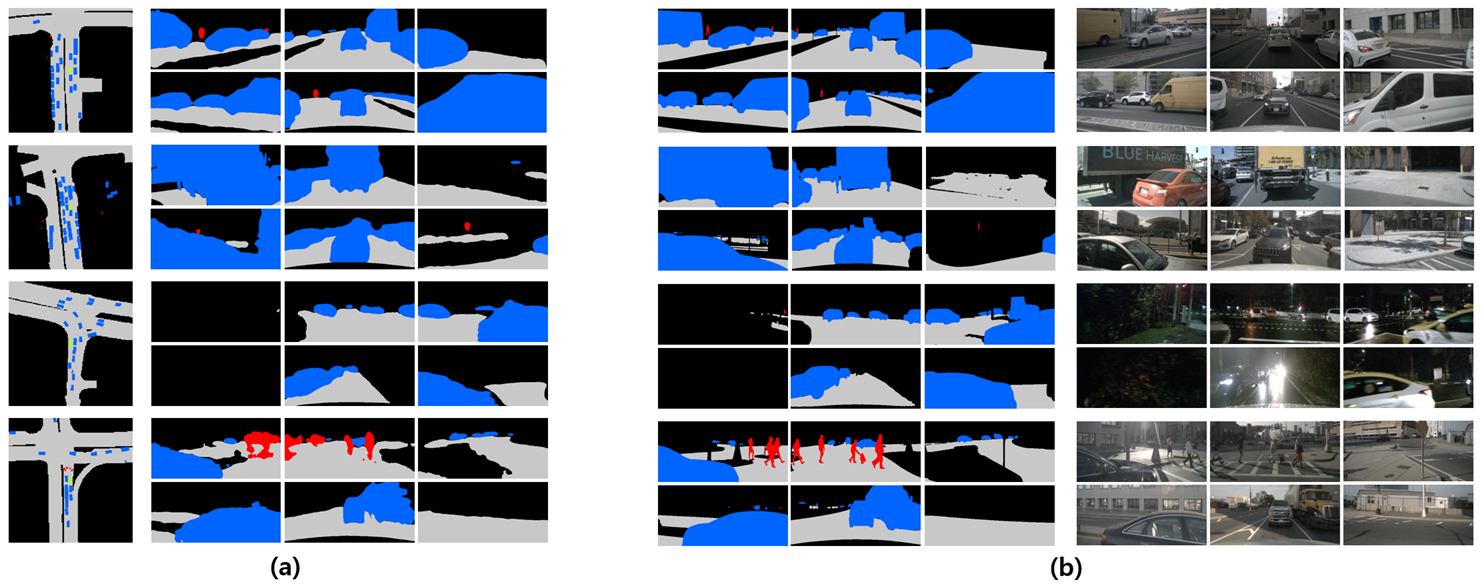}
\caption{\textbf{Examples of PV segmentation map prediction results.} (a) GT BEV maps (left) and PV map predictions from the dual-branch IVT network (right). (b) GT PV maps (left) and their corresponding multi-view images (right). \textit{Drivable area}, \textit{vehicle}, and \textit{pedestrian} are color-coded with gray, blue, and red, respectively.}
\label{fig6_supple}
\end{figure*}

Figure \ref{fig6_supple} shows PV segmentation maps predicted by the proposed dual-branch IVT network. Given that the positions of objects and road shapes in PV space are well reconstructed from the BEV maps, it can be inferred that the IVT network effectively learns the reverse mapping function. However, it fails to recover the fine outlines of \textit{vehicle} and \textit{pedestrian} instances in PV space, as these objects are represented as rectangular regions in the BEV maps. 


\begin{table}[t]
\begin{center}
\scalebox{0.8}{
\begin{tabular}{|c|c c c|c|}
\hline
\textbf{Model} & \textit{Driv.} & \textit{Veh.} & \textit{Ped.} & Avg.\\
\hline

\textbf{Single-branch} & \tcb{82.32} & \tcb{70.48} & \tcb{32.21} & \tcb{61.67} \\
\textbf{Dual-branch} & \tcr{81.82} & \tcr{68.91} & \tcr{28.27} & \tcr{59.67} \\

\hline
\end{tabular}
}
\end{center}
\caption{\textbf{PV segmentation performance of the proposed IVT network on nuScenes validation set.}}
\label{tab7_supple}
\end{table}

\begin{table}[t]
\begin{center}
\scalebox{0.65}{
\begin{tabular}{|c|c c c|c c c|c|}
\hline
\textbf{Model} & VCC & Height & Align &  \textit{Driv.} & \textit{Veh.} & \textit{Ped.} & Avg.\\
\hline
\multirow{7}{*}{\textbf{CVT}}  
  & \cellcolor{gray!20} & \cellcolor{gray!20}  &    \cellcolor{gray!20} & \cellcolor{gray!20}76.80       & \cellcolor{gray!20}31.41       & \cellcolor{gray!20}10.89       & \cellcolor{gray!20}39.70 \\

    & \cmark &  &                 & \tcb{77.20}       & \tcb{32.84}       & \tcb{12.73}       & \tcb{40.90}  \\
  & \cmark &  &                 & \tcr{76.99}       & \tcr{32.65}       & \tcr{12.00}       & \tcr{40.55}  \\

    & \cellcolor{gray!20}\cmark & \cellcolor{gray!20}\cmark & \cellcolor{gray!20} & \cellcolor{gray!20}\tcb{77.11}       & \cellcolor{gray!20}\tcb{33.62}       & \cellcolor{gray!20}\tcb{13.16}       & \cellcolor{gray!20}\tcb{41.30} \\
  & \cellcolor{gray!20}\cmark & \cellcolor{gray!20}\cmark & \cellcolor{gray!20} & \cellcolor{gray!20}\tcr{77.12}       & \cellcolor{gray!20}\tcr{33.76}       & \cellcolor{gray!20}\tcr{13.32}       & \cellcolor{gray!20}\tcr{41.40} \\

  & \cmark & \cmark & \cmark  & \tcb{77.23} & \tcb{34.14} & \tcb{13.65} & \tcb{41.67} \\  
  & \cmark & \cmark & \cmark  & \tcr{77.40} & \tcr{34.24} & \tcr{13.69} & \tcr{41.78} \\

\hline
\multirow{7}{*}{\textbf{BEVFormer}} 
  & \cellcolor{gray!20} & \cellcolor{gray!20} & \cellcolor{gray!20} & \cellcolor{gray!20}78.06       & \cellcolor{gray!20}33.23       & \cellcolor{gray!20}11.70       & \cellcolor{gray!20}41.00 \\

    & \cmark &  &                 & \tcb{78.27} & \tcb{33.77}       & \tcb{13.26}       & \tcb{41.77}  \\
  & \cmark &  &                 & \tcr{78.20} & \tcr{33.61}       & \tcr{13.32}       & \tcr{41.71}  \\

  & \cellcolor{gray!20}\cmark & \cellcolor{gray!20}\cmark & \cellcolor{gray!20} & \cellcolor{gray!20}\tcb{78.11}       & \cellcolor{gray!20}\tcb{34.35}       & \cellcolor{gray!20}\tcb{13.06}       & \cellcolor{gray!20}\tcb{41.84} \\  
  & \cellcolor{gray!20}\cmark & \cellcolor{gray!20}\cmark & \cellcolor{gray!20} & \cellcolor{gray!20}\tcr{78.11}       & \cellcolor{gray!20}\tcr{34.26}       & \cellcolor{gray!20}\tcr{13.04}       & \cellcolor{gray!20}\tcr{41.80} \\

  & \cmark & \cmark & \cmark   & \tcb{77.76} & \tcb{34.31} & \tcb{13.34} & \tcb{41.80} \\  
  & \cmark & \cmark & \cmark   & \tcr{78.20} & \tcr{34.46} & \tcr{13.39} & \tcr{42.02} \\

\hline
\end{tabular}
}
\end{center}
\caption{\textbf{Ablation study on the effectiveness of our contributions.} \textit{VCC}, \textit{Height}, and \textit{Align} refer to view cycle consistency, height prediction, and intermediate BEV feature alignment, respectively. Results highlighted in blue and red are produced using a \textcolor{blue}{single}- and \textcolor{red}{dual}-branch IVT networks, respectively.}
\label{tab8_supple}
\end{table}

Table \ref{tab7_supple} reports the PV segmentation results of the proposed IVT network with the single- and dual-branch designs. The main difference between the two designs is that the former fuses the MR feature maps generated by the CNN encoder progressively in one IVT encoder, whereas the latter first processes high and low resolution feature maps separately through dedicated IVT encoders and then fuses the resulting features in the decoding process. From now on, we refer to the single- and dual-branch designs as \textit{early} and \textit{late} fusion architectures, respectively. The table shows that the early fusion achieves better PV segmentation accuracy than its late fusion counterpart, leading to stronger regularization on VT models (see Table \ref{tab8_supple}, baselines with VCC only). However, when the proposed auxiliary tasks are incorporated into training, the late-fusion design provides better regularization for the baselines. We speculate that the early fusion helps PV segmentation because the network gets a unified view of spatial and semantic cues. On the other hand, the late fusion keeps high- and low-resolution streams separate for longer. When the auxiliary tasks are added, that separation gives each branch room to specialize---one can focus on fine spatial detail, the other on semantic abstraction---before combining them. This diversity in representation may regularize the learning process better. The fact that the late fusion benefits from the intermediate BEV feature alignment (referred to as \textit{Align} in the table) more than the early fusion does validates our claim.

\subsection{Performance Gain under Temporal Setting}
We apply the proposed regularization framework to BEVFormer \cite{Li_eccv22} under a temporal setting, where the model is configured to leverage information from previous frames, to see how much performance gain can be achieved by the proposed framework on top of the temporal setting. Specifically, we train the model to take $N_{p}$ previous frames in addition to the current frame as input. Since the key frames in nuScenes are sampled at 2 Hz, the model observes a $0.5 \times N_{p}$-second history. Table \ref{tab9_supple} shows the result. One noticeable finding is that the static model (BEVFormer-S) outperforms the temporal model (BEVFormer-T) on \textit{drivable area}, which can also be observed in the original paper \cite{Li_eccv22}. We speculate that, unlike \textit{vehicle} and \textit{pedestrian}, \textit{drivable area} lacks distinctive characteristics that can be matched to those in previously observed scenes. With respect to \textit{vehicle} and \textit{pedestrian}, BEVFormer-T outperforms BEVFormer-S, which is consistent with the findings in the original paper. Finally, the proposed regularization framework consistently improves BEVFormer-T for all $N_p$ values (see $\Delta_{\text{Ours}}$ in the table).

\begin{table}[t]
\begin{center}
\scalebox{0.65}{
\begin{tabular}{|c|c|c c c|c|}
\hline
\textbf{Model}        & $N_{p}$ & \textit{Driv.} & \textit{Veh.} & \textit{Ped.} & Avg.\\
\hline
\textbf{BEVFormer-S}  & 0       & 78.06          & 33.23         & 11.70         & 41.00 \\
\hline
\multirow{3}{*}{\textbf{BEVFormer-T}}  
               & 1                 & 76.55 & 33.50 & 12.36 & 40.80 \\
               & 2                 & 76.55 & 33.47 & 12.69 & 40.90 \\
               & 3                 & 75.65 & 33.72 & 12.41 & 40.59 \\
\hline
\multirow{3}{*}{$\Delta_{\text{Ours}}$ }

               & 1                 & 0.1 & 0.95 & 1.34 & 0.8 \\
               & 2                 & -0.3 & 0.93 & 1.36 & 0.67 \\
               & 3                 & 0.84 & 1.18 & 1.56 & 1.2 \\        

\hline
\end{tabular}
}
\end{center}
\caption{\textbf{Performance gain on BEVFormer with the temporal setting.} $N_{p}$ denotes the number of previous frames the models take as input. $\Delta_{\text{Ours}}$ denotes the performance gain achieved by BEVFormer-T+Ours over BEVFormer-T. }
\label{tab9_supple}
\end{table}

\begin{table}[t]
\begin{center}
\scalebox{0.65}{
\begin{tabular}{|c|c|c c c|}
\hline
\textbf{Model}       & $N_{p}$ & \textit{Driv.} & \textit{Veh.} & \textit{Ped.}\\
\hline
\textbf{BEVFormer-S} & 0       & 78.06          & 33.23         & 11.70 \\      
\hline         
\multirow{3}{*}{$\Delta_{\text{Temp}}$ }

               & 1                 & \tcb{-1.51} & \tcb{0.27} & \tcb{0.66} \\
               & 2                 & \tcb{-1.51} & \tcb{0.24} & \tcb{0.99} \\
               & 3                 & \tcb{-2.41} & \tcb{0.49} & \tcb{0.71} \\   
\hline
$\Delta_{\text{Ours}}$ & 0                 & \tcr{0.14} & \tcr{1.23} & \tcr{1.69} \\   
     
\hline
\end{tabular}
}
\end{center}
\caption{\textbf{Comparison with Temporal Information Aggregation.} $N_{p}$ denotes the number of previous frames the models take as input. The values in blue are the gains achieved by BEVFormer-T over BEVFormer-S. The values in red are the gains achieved by BEVFormer-S+Ours over BEVFormer-S.}
\label{tab10_supple}
\end{table}

\subsection{Comparison with Temporal Setting}
In Table \ref{tab10_supple}, we further compare the static model trained under the proposed regularization framework (BEVFormer-S+Ours) and the temporal model (BEVFormer-T) to show the effectiveness of the proposed framework. The values in red and blue are the performance gain over BEVFormer-S, respectively. It is shown that the performance gain achieved by the proposed framework (values in red) surpasses that achieved by the temporal information (values in blue) even though the proposed framework does not cause an increase in computational complexity and network size, whereas the temporal setting does. Finally, it is worth mentioning that comparing the two methods on \textit{drivable area} is unfair because, as we discovered in Tab. \ref{tab9_supple}, the temporal model performs poorly on the category.

\subsection{Training Cost} 
The table below summarizes the training-time resource consumption. All models were trained using four RTX 4090 GPUs (24GB) with batch size 2 for 30 epochs. Let $L_q$, $L_k$, and $L_k^{\text{def}}$ denote the numbers of queries, keys, and sampling points in deformable attention, respectively. The memory complexities of CVT and BEVFormer are $\mathcal{O}(L_q L_k)$ and $\mathcal{O}(L_q L_k^{\text{def}})$, respectively, where generally $L_k^{\text{def}} \ll L_k$.
\begin{center}
\centering
\scalebox{0.5}{
\begin{tabular}{|c|c|c|c|c|}
\hline
\textbf{Model} & \textbf{Sec.$/$Iter} & \textbf{Train. Time(Hrs)} & \textbf{GPU mem.$/$Batch(GB)} & \textbf{$\#$ Param.(M)} \\
\hline
CVT                 & 0.2  & 6 & 6.14 & 4.39 \\
CVT+IVT-Dual        & 0.41 & 12 & 8.18 & 26.74 \\
\hline
BEVFormer           & 0.24 & 7 & 3.58 & 31.17 \\
BEVFormer+IVT-Dual  & 0.43 & 12.5 & 5.87 & 53.93 \\
\hline
\end{tabular}
}
\end{center}
The proposed framework results in roughly a $2\times$ increase in training time and a $1.6\times$ increase in GPU memory usage. However, it introduces no inference-time overhead and converges within a similar number of epochs. Future work will focus on optimizing the training framework to reduce training time and GPU mem. consumption.

\subsection{Pseudo-Label (PL) Quality} 
Since nuScenes partially provides GT PV seg. labels for multi-view images, we train Mask2Former on the available labels and use it to generate PLs. As shown in Fig.~6 of the supple. material, they are less accurate than GT labels but \textit{remain sufficiently reliable for training the IVT net. for the regularization}, as validated by the consistent improvements reported in Tab. 1$\&$2. To further evaluate the robustness of the proposed framework to PL quality, we vary both the PL generator (Mask2Former vs. UNet) and its training budget.
\begin{center}
\centering
\scalebox{0.7}{
\begin{tabular}{|c|c|c|c|}
\hline
\textbf{Exp. ID} & \textbf{PL generator} & \# \textbf{Training epochs} & \textbf{PV seg. mIoU} \\
\hline
(1) & Mask2Former & 10 & 78.97 \\
\hline
(2) & Mask2Former & 5 & 76.34 \\
\hline
(3) & UNet & 10 & 67.67 \\
\hline
\end{tabular}
}
\end{center}

\begin{center}
\centering
\scalebox{0.7}{
\begin{tabular}{|c|c|c|}
\hline
\textbf{IVT pre-trained on} & \textbf{BEV model} & \textbf{BEV seg. mIoU} \\
\hline
\multirow{2}{*}{-}
 & CVT &  39.70     \\
 & BEVFormer & 41.00 \\
\hline
\multirow{2}{*}{PLs from (1)}
 & CVT+IVT &   41.78$_{2.08\uparrow}$ \\
 & BEVFormer+Ours & 42.02$_{1.02\uparrow}$ \\
\hline
\multirow{2}{*}{PLs from (2)}
 & CVT+IVT &  41.44$_{1.74\uparrow}$ \\
 & BEVFormer+Ours & 41.94$_{0.94\uparrow}$ \\
\hline
\multirow{2}{*}{PLs from (3)}
 & CVT+IVT & 41.56$_{1.86\uparrow}$ \\
 & BEVFormer+Ours & 41.62$_{0.62\uparrow}$ \\
\hline
\end{tabular}
}
\end{center}
As shown in the top table, Mask2Former trained for 10 epochs (used in the main paper) achieves the highest PV seg. accuracy, while the remaining settings produce progressively lower-quality PLs. We then pre-train the IVT net. using PLs from each setting and apply it to train the two models. The bottom table shows that the proposed framework consistently improves the performance even when the IVT is pre-trained on lower-quality PLs, indicating that the regularization effect is robust to PL noise rather than relying on highly accurate PV supervision. The lack of full GT PV labels remains a dataset limitation, and richer supervision may further improve performance.

\begin{figure}[t]
\centering
\includegraphics[height=6.0cm]{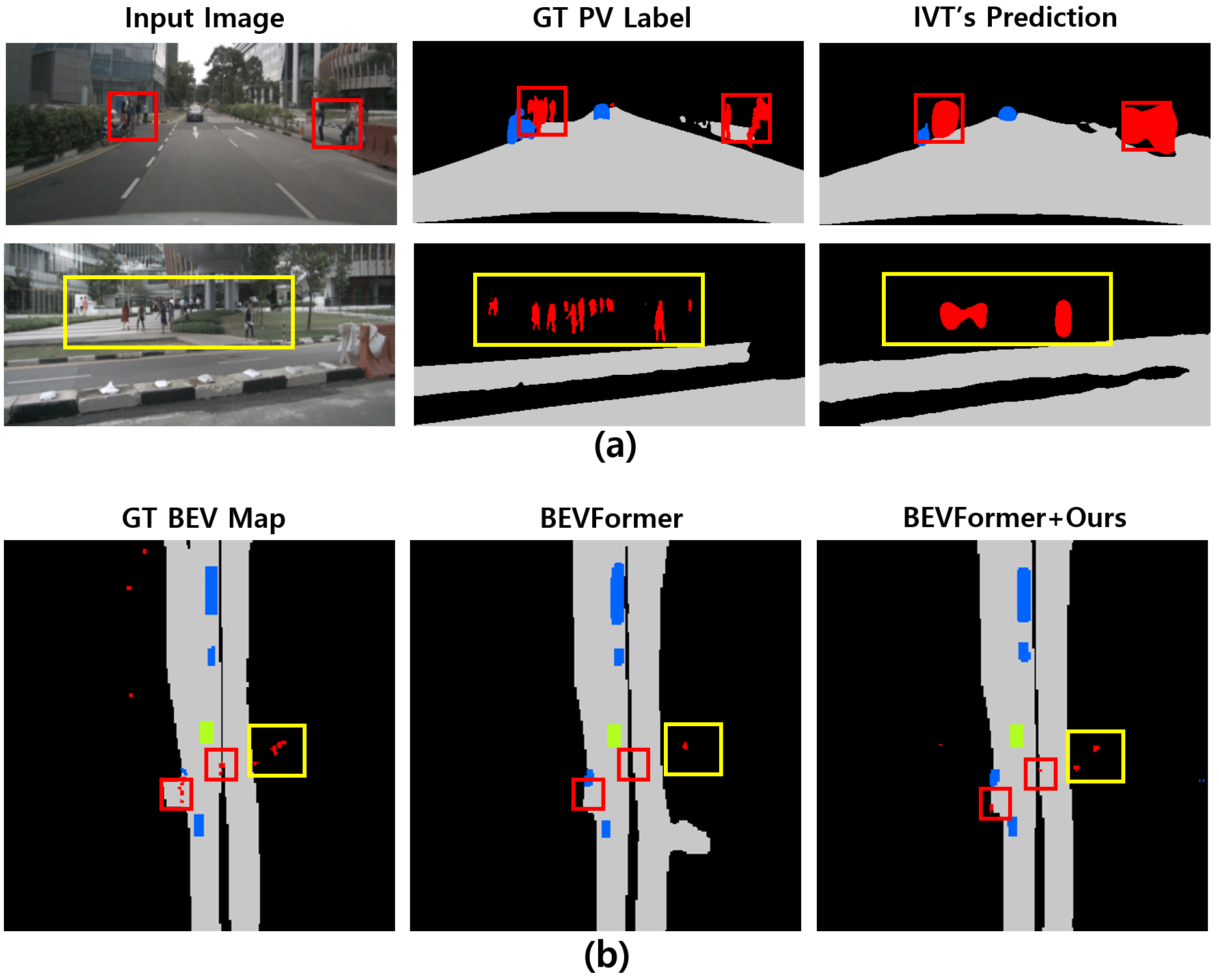}
\caption{\textbf{Prediction examples on a scene with occluded pedestrians.} (a) Input images (the first column), ground-truth PV segmentation maps (the second column), and PV segmentation maps predicted by the proposed IVT network (the third column). (b) Ground-truth BEV map (the first column), BEV map predicted by BEVFormer (the second column), and BEV map predicted by BEVFormer+Ours (the third column). The green boxes indicate the AV. Please zoom in for better visibility.}
\label{supple_fig7}
\end{figure}

\section{Additional Prediction Results}
\begin{figure*}[t]
\centering
\includegraphics[height=12.0cm, width=17.0cm]{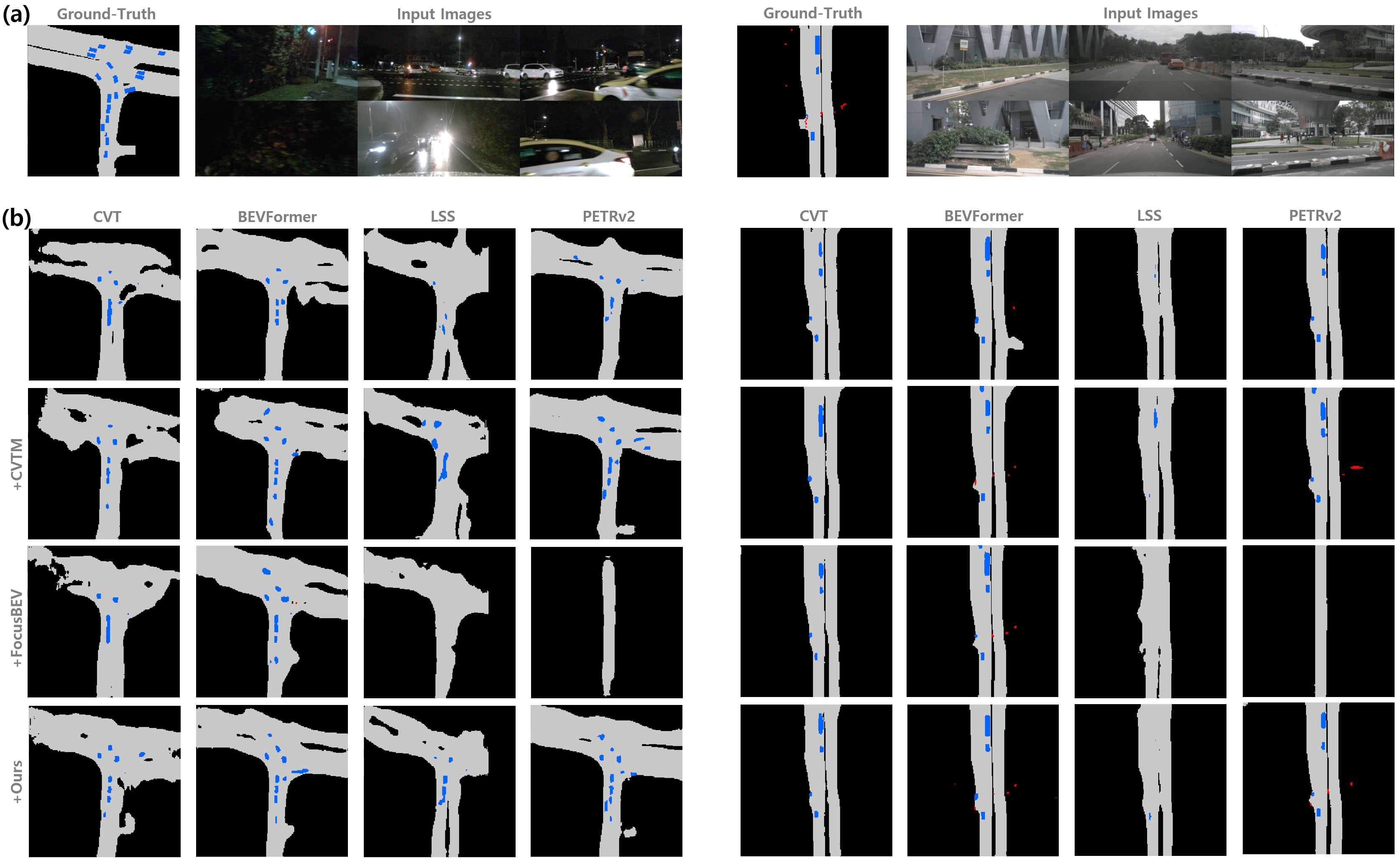}
\caption{\textbf{Prediction results.} (a) Input images and their corresponding ground-truth BEV maps, (b) BEV map prediction results. In (b), the first row shows the predictions from the four baseline models. The second, third, and fourth rows show the results when CVTM \cite{Yang_cvpr21}, FocusBEV \cite{Zhao_arxiv24}, and Ours are applied to the baseline models, respectively. \textit{Drivable area}, \textit{vehicle}, and \textit{pedestrian} are color-coded with gray, blue, and red, respectively. Please zoom in for better visibility.}
\label{supple_fig8}
\end{figure*}

\begin{figure*}[t]
\centering
\includegraphics[height=12.0cm, width=17.0cm]{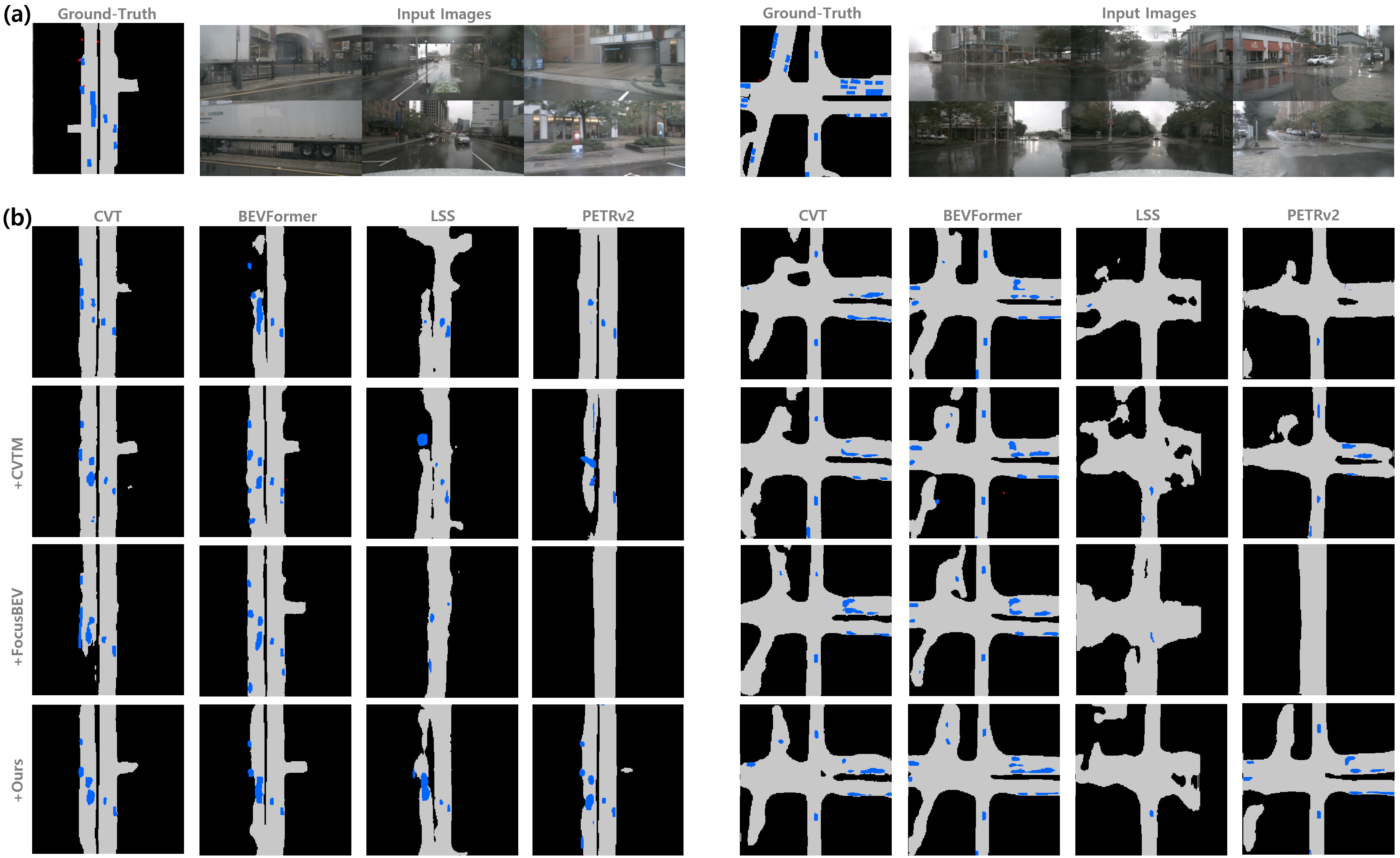}
\caption{\textbf{Prediction results.} (a) Input images and their corresponding ground-truth BEV maps, (b) BEV map prediction results. In (b), the first row shows the predictions from the four baseline models. The second, third, and fourth rows show the results when CVTM \cite{Yang_cvpr21}, FocusBEV \cite{Zhao_arxiv24}, and Ours are applied to the baseline models, respectively. \textit{Drivable area}, \textit{vehicle}, and \textit{pedestrian} are color-coded with gray, blue, and red, respectively. Please zoom in for better visibility.}
\label{supple_fig9}
\end{figure*}

Figure \ref{supple_fig7} shows the prediction results for a scene where pedestrians stand close together in the distance, making them look tiny and partially occluded. While BEVFormer misses nearly all pedestrians around the AV, BEVFormer+Ours successfully detects several of them, as highlighted by the red and yellow boxes in the figure. Highlighting the presence of the pedestrians in the input images through the learned reverse mapping helps the VT model more effectively recognize and detect the pedestrians on the BEV map. Finally, we further present the prediction results of the four baselines with or without the proposed framework, CVTM \cite{Yang_cvpr21}, FocusBEV \cite{Zhao_arxiv24} in Fig. \ref{supple_fig8} and \ref{supple_fig9}. As shown in the figure, the proposed framework improves the performance of the four baselines across all categories.

\cellcolor{gray!20}

\cellcolor{gray!20}


\end{document}